\documentclass[9.5pt,journal,compsoc]{IEEEtran}
\IEEEoverridecommandlockouts
\usepackage[colorlinks=true,linkcolor=cyan,citecolor=blue]{hyperref}
\usepackage{amsmath,amssymb,amsfonts,cite}
\usepackage{amsthm}
\usepackage{graphicx}
\usepackage{textcomp}
\usepackage{dsfont}
\usepackage{xcolor}
\usepackage{float}
\usepackage{algorithm}
\usepackage{algpseudocode, placeins}
\newtheorem{theorem}{Theorem}
\newtheorem{rem}{Remark}
\newtheorem{lemma}{Lemma}

\newtheorem{defn}{Definition}
\newtheorem{assu}{Assumption}
\newtheorem*{Remark*}{Remark}

\newcommand{\R}{\mathbb{R}}

\newcommand{\bX}{\mathbf{X}}

\newcommand{\bY}{\textbf{Y}}

\newcommand{\bE}{\mathbf{E}}
\newcommand{\bP}{\mathbf{P}}
\newcommand{\bQ}{\mathbf{Q}}
\newcommand{\bA}{\mathbf{A}}
\newcommand{\bW}{\mathbf{W}}

\newcommand{\bB}{\mathbf{B}}

\newcommand{\bM}{\mathbf{M}}

\newcommand{\bI}{\mathbf{I}}

\newcommand{\bde}{\begin{defn}}
\newcommand{\ede}{\end{defn}}

\newcommand{\Xhi}{\widehat{\bX}_M^{(i)}}
\newcommand{\Xt}{\tilde{\mathbf{X}}}
\newcommand{\Xtn}{\tilde{X}}

\newcommand{\oo}{\texttt{OmniMatch} }

\newcommand{\rr}{r_{\sigma^{(i)}(v)}}
\newcommand{\ttt}{\theta_{\sigma^{(i)}(v)}}
\newcommand{\nnn}{\nu_{\sigma^{(i)}(v)}}
\newcommand{\bb}{\vec{b}_v^{\,(i)}}
\newcommand{\rrh}{\hat{r}_{\sigma^{(i)}(v)}}

\newcommand{\bwX}{\widehat{\mathbf{X}}}

\newcommand{\ww}{Y^{(i)}_v}
\newcommand{\wwh}{\widehat{Y}^{(i)}_v}
\newcommand{\wwhj}{\widehat{Y}^{(j)}_v}
\newcommand{\wwhjw}{\widehat{Y}^{(j)}_w}

\newcommand{\eer}{\mathbin{\rotatebox[origin=c]{90}{$\Vdash$}}}
\newcommand{\neer}{\mathbin{\rotatebox[origin=c]{90}{$\nVdash$}}}

\DeclareMathOperator*{\argmin}{arg\,min}

\begin{document}

\title{Asymptotically perfect seeded graph matching without edge correlation (and applications to inference)}

\author{ Tong Qi, ~Vera Andersson, ~Peter Viechnicki, ~Vince Lyzinski
% <-this % stops a space
\IEEEcompsocitemizethanks{
\IEEEcompsocthanksitem Tong Qi, ~Vera Andersson, and Vince Lyzinski are with the Department of Mathematics at the University of Maryland, College Park. E-mail: tqi@umd.edu, veraand@umd.edu, vlyzinsk@umd.edu.
\IEEEcompsocthanksitem Peter Viechnicki led  the HLTCOE, Johns Hopkins University from 2020 to 2025. E-mail: pviechnicki2@gmail.com
 \IEEEcompsocthanksitem The authors gratefully acknowledge support from the Air Force Office of Scientific Research (AFOSR) Complex Networks award number FA9550-25-1-0128; and from the JHU HLTCOE, especially Paul McNamee and Kevin Duh.
The English-Zulu sentence pairs are taken from the Opus machine translation data portal \cite{tiedemann-2016-opus}.
We have no competing interests to declare or conflicts of interest to report.%\protect\\
}
}

\IEEEtitleabstractindextext{%
\begin{abstract} %first sentence tweak
We present the \texttt{OmniMatch} algorithm for seeded multiple graph matching.
In the setting of $d$-dimensional Random Dot Product Graphs (RDPG), we prove that under mild assumptions, \texttt{OmniMatch} with $s$ seeds asymptotically and efficiently perfectly aligns $O(s^{\alpha})$ unseeded vertices---for $\alpha<2\wedge d/4$---across multiple networks even in the presence of no edge correlation.
We demonstrate the effectiveness of our algorithm across numerous simulations and in the context of shuffled graph hypothesis testing. 
In the shuffled testing setting, testing power is lost due to the misalignment/shuffling of vertices across graphs, 
and we demonstrate the capacity of \texttt{OmniMatch} to correct for misaligned vertices prior to testing and hence recover the lost testing power.
We further demonstrate the algorithm on a pair of data examples from connectomics and machine translation.
\end{abstract}

}

\maketitle
\IEEEdisplaynontitleabstractindextext
\IEEEpeerreviewmaketitle

\section{Introduction}\label{sec:intro}

\IEEEPARstart 
Graphs and network data structures are fundamental tools for modeling complex systems of interacting objects across disciplines including biology, neuroscience \cite{priebe2019two,finn2015functional,relion2019network}, and social sciences \cite{heard_2010,newman2002random}. 
For instance, in neuroscience connectomics \cite{roncal2013migraine} regions of interest (ROIs) in the brain can be represented as nodes, while the functional connectivity between these regions can be modeled via the edge structure of the graphs.
In the context of social networks \cite{wasserman1994social}, individuals are represented as nodes, and social relationships or interactions between individuals are denoted by edges. 
In the context of natural language processing, words or concepts can be represented as nodes \cite{marchisio-etal-2022-bilingual}, and those documents containing those nodes can be further represented as graphs where nodes are adjacent based on some properties of the documents.
These graph representations provide a versatile framework for analyzing the structural and relational properties of complex systems. 
Comparing and analyzing graphs is essential for numerous tasks, including understanding structural patterns, detecting anomalies, and hypothesis testing between two graph samples. However, a persistent challenge in graph comparison arises when node correspondences are unknown or partially scrambled \cite{vogelstein2011shuffled,saxena2025lost}, as often encountered in applications such as brain connectomics \cite{vogelstein:_fast}, anonymized social networks, or language processing \cite{marchisio-etal-2022-bilingual}.

Traditional multiple graph inference methodologies such as two-sample testing \cite{tang14:_semipar,du2023hypothesis,asta2014geometric}, 
tensor decomposition \cite{zhang2018tensor,zhang2019cross,agterberg2024statistical},  
multiple graph community detection \cite{jing2020community,pensky2019, noroozi2024sparse}, etc., 
typically rely on the assumption that node identities are perfectly aligned and observed across graphs. 
When this is not the case inferential performance can degrade \cite{saxena2025lost,vogelstein2011shuffled,lyzinski16:_infoGM}, although graph matching methods \cite{conte2004thirty, yan2016short, emmert2016fifty} can be applied to align the networks, recover the missing cross-graph node labels, and recover inferential performance.
Despite these recent advancements, statistical inference under node misalignment remains underdeveloped. 
When a subset of nodes is shuffled or unaligned between graphs, recovering the correct node correspondences becomes essential to accurately reconstruct the underlying graph structure and enable valid comparisons. 

There is a robust graph matching literature both from the algorithmic and theoretic perspectives (see Section \ref{sec:match} for further background on graph matching), though most existing theoretical work (both information theoretic bounds on de-anonymization and theoretical algorithmic guarantees)
relies on the presence of edge correlation across graphs to anchor the alignment.
In this paper, we address the problem of \textit{seeded} multiple graph matching in the setting where there is no edge correlation.
Seeds here refer to vertices across the multiple graphs whose correspondence is known \textit{a priori} \cite{mossel2020seeded,fishkind09:_seeded,lyzinski2014seeded}; multiple graph matching refers to the problem of simultaneously aligning the vertices across multiple networks.
To tackle this problem, we propose the \texttt{OmniMatch} algorithm which 
jointly embeds the seeded vertices across a collection of graphs, 
out-of-sample embeds the unseeded vertices, 
and then aligns unseeded nodes by solving an approximate defined linear assignment problem in the embedded space.
This integrated approach allows us to leverage expertise in graph embeddings \cite{athreya_survey}---see Section \ref{sec:RDPG}---to efficiently, provable asymptotically perfectly match uncorrelated graphs in the general Random Dot Product Graph setting (under mild model assumptions).
To our knowledge, this is the first result proving asymptotically perfect matching is possible in this RDPG model (in either theory or algorithmically) in the absence of edge correlation.
\texttt{OmniMatch} is then further used to perform robust statistical inference by combining node matching with hypothesis testing, enabling valid comparisons even in the presence of node shuffling.

This paper is organized as follows. We describe the Random Dot Product Graph model and joint graph embedding framework in Section \ref{sec:RDPG} and 
the graph alignment and matching problem in Section \ref{sec:aligned},
We introduce the new algorithm \texttt{OmniMatch} in Section \ref{sec:omnimatch} with our main theoretical results on its graph matching performance with multiple seeds. In Section \ref{sec:sim}, we show our simulation results on both graph matching and hypothesis testing with \texttt{OmniMatch}.
Lastly, we show \oo on a pair of real data experiments using brain data and data from machine translation. We discuss this new algorithm in Section \ref{sec:discuss}.

\vspace{3mm}
\noindent\textbf{Notation:}  The following notation will be used throughout.  For positive integer $n$, we write $[n]$ to denote $[n]=\{1,2,3,\cdots,n\}$ and we write $\vec{1}_n$ to denote the $n$-dimensional vector of all $1$'s. We write $\|\cdot \|$ as the Euclidean norm on vectors and the spectral norm on matrices, $\|\cdot \|_F$ as the Frobenius norm, $ \| \cdot \|_{2\rightarrow \infty}$ as the two-to-infinity norms. The set of $d \times d$ real orthogonal matrices is denoted by $\mathcal{O}_d$. We have $\bI_k$ to denote the $k \times k$ identity matrix. 

\subsection{Random Graph Models and Joint Graph Embeddings}
\label{sec:RDPG}

Our analysis will be couched within the context of Random Dot Product Graphs (RDPG) of \cite{young2007random}, a class of latent position graph models \cite{Hoff2002} that are particularly amenable to subsequent analysis \cite{athreya_survey}.
Note that we suspect that all results contained herein can be proven for the more general Generalized RDPG of \cite{rubin-delanchy_tang_priebe_grdpg}, though the RDPG is suitable for our present purposes.

\begin{defn}[Joint Random Dot Product Graph]
\label{def:JRDPG}
Let $F$ be a distribution on a set $\mathcal{X}\in \R^d $ satisfying $\langle x,x' \rangle \in [0,1]$ for all $x,x'\in \mathcal{X}$.
Let $X_1,X_2,\cdots,X_n\stackrel{i.i.d.}{\sim} F$,
and let $P=\bX\bX^T$, where  
$\bX=[X_1^T| X_2^T| \cdots| X_n^T]^T\in\R^{n\times d}$.
We say that the random graphs $(\bA^{(1)},\bA^{(2)},\cdots,\bA^{(m)})$ are an instantiation of a \emph{Joint
Random Dot Product Graph} model (abbreviated $\mathrm{JRDPG}$), written 
$(\bA^{(1)},\bA^{(2)},\cdots,\bA^{(m)},\bX)\sim \mathrm{JRDPG}(F,n,m)$
if the following holds:
\begin{itemize}
    \item[i.] Marginally each $(\bA^{(k)},\bX)\sim\mathrm{RDPG}(F,n)$
    ; namely, $\bA^{(k)}$ is a symmetric, hollow adjacency matrix with above-diagonal entries distributed via
\begin{align}
\label{eq:rdpg}
\mathbb{P}(\bA^{(k)}|\bX)=\prod_{i<j}(X_i^TX_j)^{A^{(k)}_{ij}}(1\!-\!X_i^TX_j)^{1-A^{(k)}_{ij}};
\end{align}
Succinctly, conditioned on $\bX$ the above diagonal entries of $\bA^{(k)}$ are independent Bernoulli random variables with success probabilities provided by the corresponding above diagonal entries in $P$.
\item[ii.] Conditioned on $\bX$, the $\bA^{(k)}$'s are independent.
\end{itemize} 
\end{defn}
\noindent Note that the assumption of independence of the $\bA^{(k)}$'s in Definition \ref{def:JRDPG} is adopted here to emphasize the lack of edge correlation across the graphs we will be aligning with \texttt{OmniMatch}.  
Adding edge correlation across the $\bA^{(k)}$ so as to be suitable for much existing matching theory  \cite{wu2021settling,cullina2016improved,cullina2017exact,lyzinski16:_infoGM} is easily achieved; see \cite{pantazis2022importance} for detail.

One of the key inference tasks in the RDPG framework is estimating the latent position matrix $\bX$.  To this end, spectral methods (in particular the Adjacency Spectral Embedding of \cite{SusTanFisPri2012}) have proven particularly useful.
Spectral methods provide suitable estimates $\widehat \bX$ of $\bX$ that are then amenable to subsequent (more) classical inference methods such as clustering \cite{SusTanFisPri2012,lyzinski13:_perfec,rohe2011spectral}, testing \cite{tang14:_nonpar,tang14:_semipar,du2023hypothesis}, classification \cite{tang2012universally}, etc; for a survey of work done in this area, see \cite{athreya_survey}.
The Adjacency Spectral Embedding is defined as follows.
\begin{defn}[Adjacency Spectral Embedding (ASE)] Let $d\geq 1$ be a positive integer.  
The $d$-dimensional \emph{Adjacency Spectral Embedding} of a graph $\bA$ into $\R^d$, denoted by $\mathrm{ASE}(\bA, d)$, is defined to be $\widehat \bX=U_{A}S_{A}^{1/2}$, where 
$$|\bA| = (\bA^{\top} \bA)^{1/2}=\left[U_A\,|\,\widetilde U_A\right]\left[S_A
\oplus \widetilde S_A\right]\left[U_A\,|\,\widetilde U_A\right]^T,$$ 
is the spectral decomposition of $|\bA|$, $S_A\in\mathbb{R}^{d\times d}$ is the diagonal matrix with the $d$ largest eigenvalues of $|\bA|$, and $U_A\in\mathbb{R}^{n\times d}$ the corresponding matrix of the $d$-largest eigenvectors.
\end{defn}

One issue when estimating the latent positions $\bX$ in the RDPG model is the inherent rotational nonidentifiability associated with the model \cite{agterberg2020two}:  If $\mathbf{Y}=\bX\bW$ for a rotation matrix $\bW\in\R^{d\times d}$, then 
$\mathbb{P}(\bA|\bX)=\mathbb{P}(\bA|\mathbf{Y})$.
As such, it is only possible to recover $\bX$ up to a rotation.
For single graph inference, this amounts to a requirement that the inference task be rotationally invariant.
For joint graph inference that rely on graph embeddings, this can be problematic.
If the embeddings are done separately, then an additional step (e.g., Procrustes rotation) must be taken to align the graphs in embedding space.
The alternative is to construct a joint embedding that simultaneously constructs an aligned embedding for each graph (see, e.g., \cite{levin2017central,arroyo2021COSIE,wang2019joint,zheng2022limit,zhang2018tensor}).
Note that the trade-off for not bypassing the need to align is that additional correlation across graphs is often induced in the embedding space by the joint embedding procedure \cite{pantazis2022importance}.

In the sequel, we will be making use of the \textit{Omnibus Joint Embedding Procedure} (OMNI) of \cite{levin2017central}.
In OMNI, we are given a collection of $n$-vertex graphs $(\bA^{(k)})_{k=1}^m$, the block matrix $\bM\in\R^{mn\times mn}$ is constructed, where the $i,j$-th block of $\bM$ is given by $(\bA^{(i)}+\bA^{(j)})/2$.
The OMNI embedding of $(\bA^{(k)})_{k=1}^m$ into $\R^d$ is then given by 
$$\widehat \bX_M=\text{OMNI}(\bA^{(1)},\bA^{(2)},\cdots,\bA^{(m)})=\text{ASE}(\bM,d).$$
In the JRDPG setting outlined above, the OMNI embedding provides a jointly consistent estimator of the true latent position matrix $\bX$.  Namely, we have
\begin{lemma}[Lemma 1 in \cite{levin2017central}]
\label{lem:2toinf}
Let $F$ be a distribution on a set $\mathcal{X}\in \R^d $ satisfying $\langle x,x' \rangle \in [0,1]$ for all $x,x'\in \mathcal{X}$.
Let $(\bA^{(1)},\bA^{(2)},\cdots,\bA^{(m)},\bX)\sim \mathrm{JRDPG}(F,n,m)$, and assume that if $\bY\sim F$, then $\mathbb{E}(\bY\bY^T)$ is rank $d$.
Then there exist constants $C,C'>0$ and integer $\mathfrak{n}$ (depending only on $F$ and $m$) and an orthogonal matrix $\bW$ such that for $n\geq \mathfrak{n}$, 
\begin{align}
\label{eq:2inf}\|\widehat{\bX}_M-(\vec{1}_m\otimes{\bX})\bW\|_{2\rightarrow\infty}\leq \frac{Cm^{1/2}\log(mn)}{\sqrt{n}}
\end{align}
with probability $1-C'n^{-2}$.
\end{lemma}
\noindent Moreover, under mild assumptions it is shown in \cite{levin2017central} that the residual estimation errors of $\widehat{\bX}_M$ (suitably rotated) estimating $\vec{1}_m\otimes{\bX}$ are row-wise asymptotically normal, providing a central-limit-theorem-like result for the estimation error.

\subsection{Alignment and matching}
\label{sec:aligned}

An assumption that is made in most joint embedding (and joint inference) methodologies is that the graphs are \textit{a priori} vertex-aligned; i.e., for each $i\in[n]$, vertex $i$ represents the same entity in each of the $m$ graphs;
notable exceptions to this assumption include \cite{kolaczyk2020averages,josephs2021network,saxena2025lost}.
In situations where this is not the case, graph matching methods can be used to recover the true alignment before further inference is pursued.
Graph matching is a broad term for the large collection of methods that are dedicated to finding an optimal alignment between the vertex sets of two (or more) networks; see the survey papers \cite{pan2010survey,zou2020survey,conte2004thirty} for a review of the field.
At its simplest, the problem of matching two $n$-vertex graphs can be cast as a special case of the quadratic assignment problem:  Given graphs $\bA$ and $\mathbf{B}$, the graph matching problem seeks to find all elements of $\text{argmin}_{P\in\Pi_n}\|\bA P-P\mathbf{B}\|_F$ where $\Pi_n$ is the set of $n\times n$ permutation matrices.
It is worth noting that if $\bA$ and $\mathbf{B}$ are allowed to be weighted, directed, and loopy (i.e., with self-edges), then this is equivalent to the NP-hard quadratic assignment problem.

Due to the practical utility of graph matching and its computational complexity, there are a huge number of graph matching algorithms in the literature.
Of note here are algorithms that provide high probability matching guarantees under different levels of network model complexity.
Examples include recent work developing efficient methods for asymptotically perfect (and near-perfect) matching in correlated Erd\H os-R\'enyi graphs (e.g., both seeded \cite{mossel2020seeded,JMLR:v15:lyzinski14a} and unseeded \cite{mao2023random,barak2019nearly,fan2020spectral}).
These algorithms often rely on \textit{edge} correlation to be present across the graph pair:  Edges within each graph are independent, and edges across graphs are independent with the exception that for each pair of vertices $\{u,v\}$ (where for a graph $\bA$, $E(\bA)$ denotes the set of edges in $\bA$):
\begin{equation}
\label{eq:core}
\text{corr}\left(\mathds{1}\left\{ \{u,v\}\!\in\! E(\bA)\right\}, 
\mathds{1}\left\{ \{u,v\}\in E(\mathbf{B})\right\}\right)=\rho_e\!>\!0.
\end{equation}
This correlation is necessary in the Erd\H os-R\'enyi and Stochastic Blockmodel settings (see \cite{wu2021settling,cullina2016improved,racz2021correlated,racz2023matching}) as in the Erd\H os-R\'enyi model all vertices are stochastically equivalent (and all vertices in each block are stochastically equivalent in the blockmodel setting).
Indeed, in Erd\H os-R\'enyi graphs, if $\bA$ and $\mathbf{B}$ are independent, then the probability 
\begin{equation}
\label{eq:mof}
    \mathbb{P}(Q\in \text{argmin}_{P\in\Pi_n}\|\bA P-P\mathbf{B}\|_F)
\end{equation} is the same for all $Q\in\Pi_n$.

When edge correlation is not present, matching must rely on structural correlation across graphs for alignment.
We note here the work in \cite{zhang2018consistent,zhang2018unseeded} which establishes algorithms for consistent ($o(n)$ error for the graphon matching problem) matching across a pair of unseeded, uncorrelated latent position networks. 
The work in \cite{zhang2018consistent,zhang2018unseeded} utilizes the common structure across latent spaces to match graphs. 
A similar thrust of work considers graph matching in random geometric graphs (see, for example, \cite{wang2022random}), in which the signal in the geometry is used to match graphs without edge correlation, though
our algorithmic approach and model is different than the work contained therein (\cite{wang2022random}, in particular, considered random complete dot-product graphs).

In \cite{fishkind2019alignment,fishkind2021phantom}, this structural correlation is termed the \emph{heterogeneity correlation}, which combines with edge correlation to define the \emph{total correlation} across graphs.
For graphs $\bA$ and $\mathbf{B}$ and permutation $P$, define
$\Delta(\bA,\mathbf{B},P)=\|\bA P-P\mathbf{B}  \|_F^2/2.$
The alignment strength \cite{fishkind2019alignment} between $\bA$ and $\mathbf{B}$ under permutation $P$ is then defined to be (where $\delta_A=|E(\bA)|/\binom{n}{2}$ is the edge density of a graph)
\begin{align*}
\mathfrak{str}(\bA,\mathbf{B},P)&=\frac{\Delta(\bA,\mathbf{B},P)}{\frac{1}{n!}\sum_{Q\in\Pi_n}\Delta(\bA,\mathbf{B},Q)}\\
&=\frac{\Delta(\bA,\mathbf{B},P)/\binom{n}{2}}{\delta_A(1-\delta_B)+\delta_B(1-\delta_A)}
\end{align*}
Alignment strength provides what its name suggests, a measure of the strength of the alignment across a pair of graphs.
In \cite{fishkind2019alignment,fishkind2021phantom} it is empirically demonstrated that matching algorithms (there the SGM algorithm of \cite{fishkind09:_seeded}) perform better and faster when aligning graphs with a higher alignment strength across them.
They further note that if $\bA$ and $\mathbf{B}$ are identically distributed with latent alignment $P$, then the alignment strength is asymptotically almost surely equivalent to the total correlation $\rho_T$ across graphs; i.e., $\mathfrak{str}(\bA,\mathbf{B},P)-\rho_T\stackrel{a.s.}{\rightarrow}0$. 
Here, $\rho_T$ is defined via 
$\rho_T=1-(1-\rho_e)(1-\rho_h)$ where $\rho_e$ is the edge-wise correlation defined in Eq. \ref{eq:core}; and $\rho_h$ is the heterogeneity correlation defined via 
    $\rho_h=\sigma^2/(\mu(1-\mu))$,
    where $\mathbf{P}=\mathbb{E}\bA=[p_{uv}]$, $\mu=\sum_{\{u,v\}
    % \in\binom{V}{2}
    }p_{uv}/\binom{n}{2}$ and
    $\sigma^2:=\sum_{\{u,v\}
    % \in \binom{V}{2}
    }(p_{uv}-\mu)^2/\binom{n}{2}.$ Just as large values of alignment strength lead to better matching outcomes empirically, so too does larger total correlation.  Of particular note is that total correlation can be (relatively) large even in the case where $\rho_e=0$, as long as there is sufficient heterogeneity correlation.

\section{OmniMatch}
\label{sec:omnimatch}

Our proposed matching methodology is designed for the setting where there is no edge correlation, significant heterogeneity correlation, and available seeded vertices.
Here, \emph{seeded vertices} refer to those vertices whose correspondence across graphs is known \textit{a priori.}
This amounts to optimizing over a restricted subspace of $\Pi_n$ in which diagonal elements (those corresponding to the seeded vertices) of the unknown permutation are fixed to be 1.
Notably, the \texttt{OmniMatch} algorithm can be used to align the unseeded vertices across any number $m$ of networks.

Before stating the algorithm, we first introduce the problem set-up.
Consider $m$ vertex aligned graphs $({\bA}^{(1)},{\bA}^{(2)},\cdots,{\bA}^{(m)})$ on $n=s+u$ vertices, though rather than observing these graphs, we observe networks where a portion of the vertices are potentially shuffled.
Here, suppose that the first $s$ vertices are aligned across the $m$ graphs and consider the remaining $u$ vertices in each graph as having unknown correspondence across the collection; denote the aligned vertices via $\mathcal{S}$ and the set of potentially shuffled vertices via $\mathcal{U}$.
We shall model this missing correspondence via each graph being observed as 
$$\bB^{(i)}=(I_s\oplus \bQ^{(i)})\bA^{(i)}(I_s\oplus \bQ^{(i)})^T,$$
and let the permutation associated with the permutation matrix $\bQ^{(i)}\in\R^{u\times u}$ be denoted $\sigma^{(i)}$. 
The \texttt{OmniMatch} algorithm proceeds as follows (see Algorithm \ref{alg:omni} for \texttt{OmniMatch} pseudocode).
\begin{itemize}
\item[i.] Let the induced subgraph of $\bB^{(i)}$ on the $s$ aligned vertices in $\mathcal{S}$ be denoted $\tilde{\bA}^{(i)}$ with corresponding latent positions $\tilde{\bX}$. 
\item[ii.] Embed the aligned vertices via $\widehat{\bX}_M=\text{OMNI}(\tilde{\bA}^{(1)},\tilde{\bA}^{(2)},\cdots,\tilde{\bA}^{(m)},d)$.  Let $\widehat{\bX}_M^{(i)}$ denote the portion of $\widehat{\bX}_M$ corresponding to graph $i$ (i.e., rows $[(i-1)*s+1]:[i*s]$).
If the embedding dimension is unknown, it can be estimated via a number of available methods (e.g., 
\cite{chen2021estimating}); we estimate the dimension via elbow-analysis of the SCREE plot based on \cite{chatterjee2014matrix,zhu2006automatic}.
The complexity of this step is $O(m^2s^2d)$ for the truncated SVD \cite{feng2018fast}.
\item[iii.] 
For each $i$, let the decomposition of $\mathbf{B}^{(i)}$ be given by
$$\mathbf{B}^{(i)}=\begin{bmatrix}\tilde{\mathbf{A}}^{(i)} & \mathbf{B}^{(i)}_{12}\\\mathbf{B}^{(i)}_{21}&\mathbf{B}^{(i)}_{22} \end{bmatrix},$$
where $\mathbf{B}^{(i)}_{22}$ is the adjacency among the unseeded vertices.  Given $\bwX^{(i)}$ from step ii., we can estimate the remaining latent positions (of the unseeded vertices) by minimizing the objective 
\begin{align}
f(\mathbf{Y}^{(i)})=&\|\mathbf{B}^{(i)}_{12}\!-\!\bwX^{(i)}_M(\mathbf{Y}^{(i)})^T\|_F^2\notag\\
&+\lambda \|\mathbf{B}^{(i)}_{22}\!-\!\mathbf{Y}^{(i)}(\mathbf{Y}^{(i)})^T\|_F^2\label{eq:fobj}
\end{align}
over $\mathbf{Y}^{(i)}$, where the final ASE is then $\mathbf{Z}^{(i)}=\big[ (\bwX^{(i)})^T | (\mathbf{Y}^{(i)})^T\big]^T$.  
Note that the the Least-Squares Out-Of-Sample (LLS OOS) embedding heuristic defined in \cite{levin2021limit} is obtained by setting $\lambda=0$; in this $\lambda=0$ case, the complexity of the OOS embedding is $O(msud^2)$ \cite{levin2018out}.  
When $\lambda > 0$, the gradient of $f(\mathbf{Y}^{(i)})$ can be computed via
\begin{align*}
\nabla_{\mathbf{Y}^{(i)}} f(\mathbf{Y}^{(i)})=&2\mathbf{Y}^{(i)}(\bwX^{(i)})^T\bwX^{(i)}-2(\mathbf{B}^{(i)}_{12})^T\bwX^{(i)}\\
&+4\lambda(\mathbf{Y}^{(i)}(\mathbf{Y}^{(i)})^T\mathbf{Y}^{(i)}-\mathbf{B}^{(i)}_{22}\mathbf{Y}^{(i)}). 
\end{align*}
When $\lambda>0$, each gradient computation has complexity (assuming $s,u\gg d$), $O(sud+u^2d)$; given a bounded number of gradient steps, the OOS would have complexity $O(mu^2d+msud)$.
While an exact solution to Eq. \ref{eq:fobj} is not immediately apparent, we can estimate the optimal $\mathbf{Y}^{(i)}$ via gradient descent, without discarding away the bulk of the matrix as in the classical OOS setting (i.e., while still using $\mathbf{B}^{(i)}_{22}$); call this obtained solution $\widehat{\mathbf{Y}}^{(i)}$.

\item[iv.] For each $i,j\in[m]$, compute the cost matrices $\mathbf{C}^{(i,j)}\in\mathbb{R}^{u\times u}$ where 
$$\mathbf{C}^{(i,j)}_{v,w}=\|\wwh-\wwhjw\|^2.$$
\item[v.] For each $i,j\in[m]$, compute the matching of the unaligned vertices in $\bB^{(i)}$ to the unaligned vertices in $\bB^{(j)}$ by finding an element of
$$\text{argmin}_{\bQ\in\Pi_u}\text{tr}(\mathbf{C}^{(i,j)}\bQ);$$
Solve the multidimensional assignment problem (MLAP see, for example, \cite{nguyen2014solving}) with cost matrices $\{\mathbf{C}^{(i,j)}\}$ to find an assignment of labels to the unlabeled vertices across all $m$ graphs.
Let the set of vertices with unknown labels in $\bB^{(i)}$ be denoted $\mathcal{U}_{\bB^{(i)}}$, and for each $\vec v=[v_i]\in \mathcal{U}_{\bB^{(1)}}\times \mathcal{U}_{\bB^{(2)}}\times\cdots \mathcal{U}_{\bB^{(m)}}$ define the cost 
$
\mathcal{C}(\vec v)=\sum_{\{i,j\}\in\binom{m}{2}}\mathbf{C}\mathbf{C}^{(i,j)}_{v_i,v_j}.
$
We then seek permutations $\pi_1,\cdots,\pi_{m-1}$ to minimize the following cost:
$\sum_{v\in \mathcal{U}_{\bB^{(1)}}}\mathcal{C}(v,\pi _{1}(v),\ldots ,\pi _{m-1}(v)).$
This could be done with an off-the-shelf MLAP solver (such as in \cite{nguyen2014solving}); as an alternative, an anchor graph can be chosen and all other graphs aligned to the anchor pairwise.
Solving the multidimensional assignment problem is NP-hard in general, though when $m=2$ this amounts to solving a simple linear assignment problem which can be solved in $O(u^3)$ time (using, for example, the Hungarian algorithm of \cite{hungarian}).
Moreover, in the theoretical analysis below we can solve pairwise linear assignment problems 
% \begin{align}
\begin{equation}
    \label{eq:lap}
\text{argmin}_{\bQ\in\Pi_u}\text{tr}(\mathbf{C}^{(i,j)}\bQ)
\end{equation}
% \end{align}
and the composition of these assignments will, with probability going to 1, be internally consistent (i.e., if $v\in \mathcal{U}_{\bB^{(j)}}$ is mapped to $w\in \mathcal{U}_{\bB^{(\ell)}}$ and $w\in \mathcal{U}_{\bB^{(\ell)}}$ is mapped to $z\in \mathcal{U}_{\bB^{(h)}}$, then $v\in \mathcal{U}_{\bB^{(j)}}$ will be mapped to $z\in \mathcal{U}_{\bB^{(h)}}$) and will provide the correct labels across all graphs; this pairwise consistency yields an MLAP approximate solution with complexity $O(mu^3)$ by picking an arbitrary reference graph and aligning everything to it.
\end{itemize}

\begin{algorithm}[t!]
\caption{\texttt{OmniMatch} for multiple seeded graph matching}\label{alg:omni}
\begin{algorithmic}
\Require $m$ graphs $({\bB}^{(1)},{\bB}^{(2)},\cdots,{\bB}^{(m)})$ each with $n=s+u$ vertices; Seeded vertex set $S=\{1,2,3,\cdots,s\}$ assumed to be the same vertices in each graph; Embedding dimension $d$
\Ensure Alignment of unlabeled vertices across $m$ graphs
\State\textbf{[i.]} For each $i\in[m]$, let $\tilde{\bA}^{(i)}=\bB^{(i)}[1:s,1:s]$;
\State\textbf{[ii.]} Embed the aligned vertices via $\widehat{\bX}_M=\text{OMNI}(\tilde{\bA}^{(1)},\tilde{\bA}^{(2)},\cdots,\tilde{\bA}^{(m)},d)$;
\State\textbf{[iii.]} For each $v\in\mathcal{U}=V\setminus S$ and $i\in[m]$, if $\lambda>0$, use gradient descent applied to Eq. \ref{eq:fobj} to obtain the
OOS embedding $\widehat{\mathbf{Y}}^{(i)}$ of the unseeded vertices in $\bB^{(i)}$; if $\lambda=0$, set 
$\widehat{\mathbf{Y}}^{(i)}=(\mathbf{B}^{(i)}_{12})^T\bwX^{(i)}((\bwX^{(i)})^T\bwX^{(i)})^{-1}$;
\State\textbf{[iv.]} For each $i,j\in[m]$, compute the cost matrices $\mathbf{C}^{(i,j)}\in\mathbb{R}^{u\times u}$ where we define 
$\mathbf{C}^{(i,j)}_{v,w}=\|\wwh-\wwhjw\|^2;$
\State\textbf{[v.]} Solve the multidimensional assignment problem with cost matrices $\{\mathbf{C}^{(i,j)}\}$ to find an assignment of labels to the unlabeled vertices across all $m$ graphs;
\end{algorithmic}
\end{algorithm}

\noindent Note that when the graphs are of different orders (i.e., differing numbers of vertices) or only partially overlap, \texttt{OmniMatch} can still be applied by allowing vertices to be unassigned in the final LAP step.

\begin{rem}
\label{rem:remlambda}
\emph{
In step iii.\@ of the \oo 
algorithm, the weight $\lambda$ appears as a hyperparameter.
While we have not proven an optimal value, in experiments $\lambda=0.5$ (or $\lambda=0.2$) seem to work well in practice; our default will be $\lambda=0.5$ unless otherwise specified.
Considering the objective function in Eq. \ref{eq:fobj}, this is reasonable as it matches the coefficients of each term in the full objective function $\|\bB^{(i)}-\mathbf{Z}^{(i)}(\mathbf{Z}^{(i)})^T\|_F^2/2$; for an empirical exploration of the effect of $\lambda$, see Figure \ref{fig:dim}.
Note that the $\lambda=0$ case is of particular importance, as this would be needed in the setting where the edges among the non-seeded vertices is unknown (e.g., as in snowball sampled graphs \cite{rohe2015network}).
}
\end{rem}
\begin{rem}
\label{rem:remproc}
\emph{
A natural alternative to \oo is the following Procrustes-based method (called \texttt{ProcMatch} in experiments): Embed the graph collection separately, use orthogonal Procrustes to align the embedded seed vertices, and align the unseeded vertices as in steps \emph{iv.} and \emph{v.} in \oo.
When $m>2$, the orthogonal Procrustes step can be replaced via Generalized Procrustes Analysis \cite{gower_procrustes}.
The chief advantage of the \oo approach is its theoretical tractability, as the theoretical properties of the Procrustes alignment step are nuanced.
Curiously, when $\lambda=0.5$, \oo and \texttt{ProcMatch} seem to produce notably similar results in pair-wise matching simulations; further investigation of this is warranted and is the subject of current research.}
\end{rem}

\subsection{Theoretical analysis}
\label{sec:theory}

Herein, we provide a theoretical analysis of the \texttt{OmniMatch} algorithm in the setting where $\lambda=0$, and prove our main algorithmic consistency result.
While we suspect that a similar (or perhaps stronger) consistency result is possible in the case of $\lambda>0$, this is more difficult to establish, and we plan to pursue it in future work. 
Let $(\bA^{(1)},\bA^{(2)},\cdots,\bA^{(m)},\bX)\sim\text{JRDPG}(F,n=s+u,m)$ where $F$ is a distribution in $\mathbb{R}^d$ satisfying the following Assumption.
\begin{assu}
\label{ass:ass1}
We will assume the inner product distribution $F$ further satisfies
\begin{itemize}
\item[i.] For all $x,y\in\text{supp}(F)$, there exists a constant $\delta>0$ such that $\delta\leq x^T y\leq 1-\delta$ (we denote this property as the $\delta$-inner product property);
\item[ii.] If $\bY\sim F$, then $\mathbb{E}(\bY\bY^T)$ is rank $d$;
\item[iii.] There exists a constant $c_d>0$ (depending on the latent dimension $d$, which is fixed) such that for any $x\in\text{supp}(F)$, we have that $\mathbb{P}_F(X\in \mathrm{B}_r(x))\leq c_d r^d$; (this is not so onerous as the volume of the ball of radius $r$ about $x$ is proportional to $r^d$)
\end{itemize}
\end{assu}

\noindent Theorem 7 in \cite{levin2021limit}, adapted here to account for the factor of $m$ appearing in the Omnibus two--to--infinity bound, yields that for the same $\bW$ as in Eq. \ref{eq:2inf}, we have the following (where $X_v$ is the row of $\bX$ corresponding to vertex $v$).
\begin{lemma}
\label{lem:2infOOS}
With notation and Assumption \ref{ass:ass1} on $F$ as above, there exist constants $D,D'>0$, rotation matrix $\bW$ (independent of out-of-sample-vertex $v$ and graph index $i$), and integer $\mathfrak{s}>0$ (all depending only on $F$ and $m$) such that for all $s\geq \mathfrak{s}$,
\begin{align}
\label{eq:2infi}
\|\wwh-\bW^TX_v\|_{2}\leq D\frac{m^{1/2}\log(ms)}{\sqrt{s}}
\end{align}
holds with probability $1-D's^{-2}$.
\end{lemma}
\noindent 
Note that the $\bW$ appearing in Lemmas \ref{lem:2toinf} and \ref{lem:2infOOS} can be taken to be equal.

Note that the non-asymptotic form of this Lemma (deviating from the ``big-O'' version in \cite{levin2021limit}; this is needed to apply a union bound over all $v\in\mathcal{U}$ and over the $m$ graphs) does require some nontrivial work; for the proof of the result, see Appendix A.1.
Considering the bound in Eq. \ref{eq:2infi} uniformly over all $u$ unknown vertices and all $m$ graphs, we have that for all $s\geq \mathfrak{s}$
\begin{align}
\label{eq:2infu}
\max_{i\in[m],v\in\mathcal{U}}\|\wwh-\bW^TX_v\|_{2}\leq\frac{Dm^{1/2}\log(ms)}{\sqrt{s}}
\end{align}
holds with probability at least $1-D'mus^{-2}$.

We are now ready to state and prove our main result.
With notation and assumptions as above, consider the \texttt{OmniMatch} algorithm where we output the solutions to the $\binom{m}{2}$ LAP's in step [v]. as its final output.  
Our main result proves that this iteration of \texttt{OmniMatch} asymptotically almost surely perfectly aligns all unaligned vertices across all $m$ graphs (proven in Appendix A.1.2).
\begin{theorem}
\label{thm:main}
With notation and Assumption \ref{ass:ass1} as above, there exist constants $D',D''$ and integer $\mathfrak{s}''>0$ such that if $s\geq \mathfrak{s}''$, the probability that \texttt{OmniMatch} perfectly aligns the unaligned vertices across all $m$ graph simultaneously is at least (where $C$ is a constant independent of $d$)
$$1-p_{m,u,s,d}:=1-Cc_d 10^dD^d  m^{d/2}u^2\frac{\log^{2d}(ms) }{s^{d/2} }-D'mus^{-2};$$ i.e, with probability at least $1-p_{m,u,s,d}$ we have that for all $i,j\in[m]$,
$\argmin_{\bQ\in\Pi_u}\text{tr}(\mathbf{C}^{(i,j)}\bQ)=\{ \bQ^{(j)}(\bQ^{(i)})^T\}.$
\end{theorem}
\noindent Note that if $u=s^\alpha$ for $\alpha<\min(2,d/4)$ with $m$ fixed, then Theorem \ref{thm:main} provides eventual perfect performance for \texttt{OmniMatch} with probability going to 1; note that the constants in the bounding probabilities do depend on $d$, and may be very large for moderate $d$.
This asymptotically perfect matching is in spite of the fact that \emph{there is no edge correlation} across any of the graphs in our collection.
That said, the constraint on $F$ and the dimension $d$ (especially when $d$ is modestly large) do provide a modest level of \emph{total correlation} \cite{fishkind2019alignment} across the networks---due to the presence of heterogeneity correlation across graphs---and this is enough to asymptotically perfectly align the vertices across networks.
    \begin{rem}
    \emph{Note that Assumption \ref{ass:ass1} precludes discrete latent distributions (such as the Stochastic Blockmodel (SBM) \cite{Holland1983}) from the 
    theory.
    This is a feature and not a bug, as in the discrete case the lack of edge correlation makes it infeasible to perfectly match among the stochastically identical nodes sharing a common latent position.
    The assumptions precluding discrete distributions are necessary to ensure sufficient heterogeneity correlation exists across networks.  
    That said, it is immediate from our theory that \oo applied to a pair of uncorrelated SBMs would still asymptotically perfectly match the block assignments across graphs (i.e., under mild assumptions on the seed set, for all blocks $i$, all vertices from block $i$ in graph 1 would be mapped to vertices in block $i$ in graph 2).}
    \end{rem}
\begin{figure}[t!]
\begin{center}
\includegraphics[width=.48\textwidth]{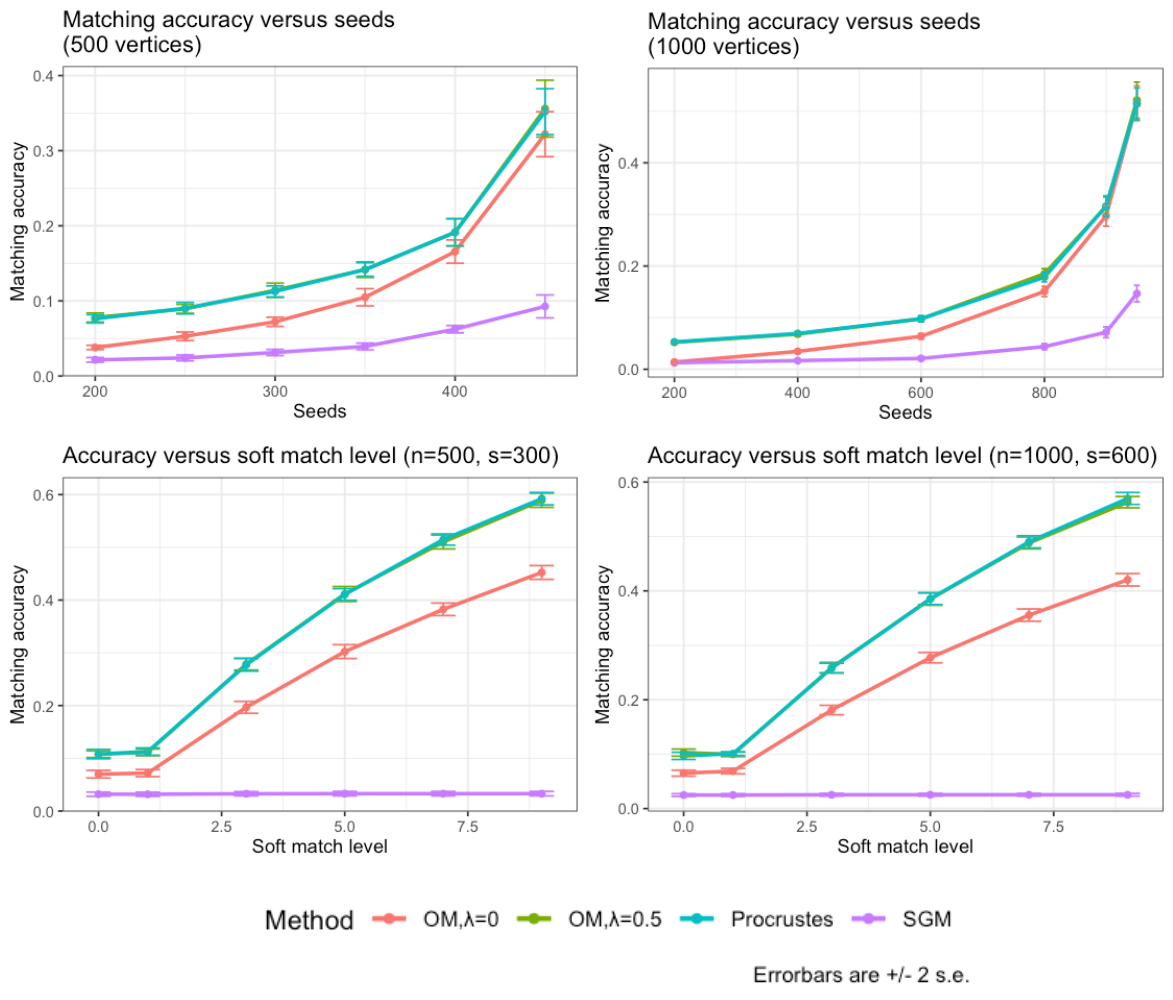}
\caption{Proportion of 
correctly unshuffled vertices for \oo, \texttt{SGM}, and \texttt{ProcMatch} versus seed count (top) and versus soft-matching level (bottom).  The left column shows matchings of graphs with $n=500$ total vertices, the right $n=1000$ vertices.
In all cases, the graphs are 2-d RDPGS; results are averaged over 30 Monte Carlo replicates.}
\label{plt:exp_prop_n500}
\end{center}
\end{figure}
    \begin{rem}
    \emph{In Theorem \ref{thm:main}, 
    the dimension $d$ acts as both blessing and curse. 
    While the effect of the seeds grows with $d$, so does the lead order constant $D''$.
    Here, $D''\propto c_d D^d 10^d$, where $D$ is the constant appearing in Lemma \ref{lem:2infOOS}, and $c_d$ the constant appearing in Assumption iii.  
    When $D$ is even modestly sized, this lead order constant can be quite large, requiring $s$ (and hence $n$) to be quite large to overcome $D''$ in theory.
    We suspect that this is somewhat an artifact of our proof technique, and we are working to reduce the size of these constants.
    In practice, we select $d$ using an automatic dimensionality selection method (locating an appropriate elbow of the SCREE plot) based on \cite{zhu2006automatic,chatterjee2014matrix}. 
    One could also employ alternative methods based on (for example)  cross-validation methods \cite{chen2021estimating}.}
    \end{rem}
\begin{rem}
    \emph{
    A natural extension would be to derive the analogous theory when $(\bA,\bX)\sim$RDPG$(F_1)$ and $(\bB,\bY)\sim$RDPG$(F_2)$ where $\bX$ is sufficiently close to $\bY$.
    If $\bX$ and $\bY$ differ by a few non-seed rows, we suspect that \oo will be robust to this noise.  If the seed sets have different latent positions (see \cite{draves2020bias} for exploration of the bias/variance here) or the differences are more global, the performance of \oo is less clear. 
    Previous results in this direction (matching across heterogeneous network pairs) \cite{het} relied entirely on the existence of edge-correlation (indeed, they first removed the differing first order structures), and it is natural to try to understand the limitations of \oo in this setting; this is the subject of ongoing research and we will not address this further in the current manuscript.
    }
\end{rem}

\section{Simulations And Real Data Experiments}
\label{sec:sim}

Herein, we show the effectiveness of \texttt{OmniMatch} for both aligning graphs and as a preprocessing tool used prior to performing multiple graph hypothesis testing.  
In this section, we also compare \oo performance to the \texttt{SGM} algorithm of \cite{fishkind09:_seeded} (as implemented in the \texttt{R} package \texttt{iGraphMatch} \cite{qiao2021igraphmatch})
and to \texttt{ProcMatch}.
For the code and real data needed to run the experiments below, we refer the reader to \url{https://github.com/tong-qii/Omnimatch} and \url{https://www.math.umd.edu/~vlyzinsk/Omnimatch}.
Note that when $\lambda>0$, unless otherwise specified, we use the following parameters for \oo: learning rate = 1e-3, tol = 1e-4, max$\_$iter = Inf (i.e., run until convergence under the provided tolerance).

\begin{figure}[t!]
\begin{center}
\includegraphics[width=.48\textwidth]{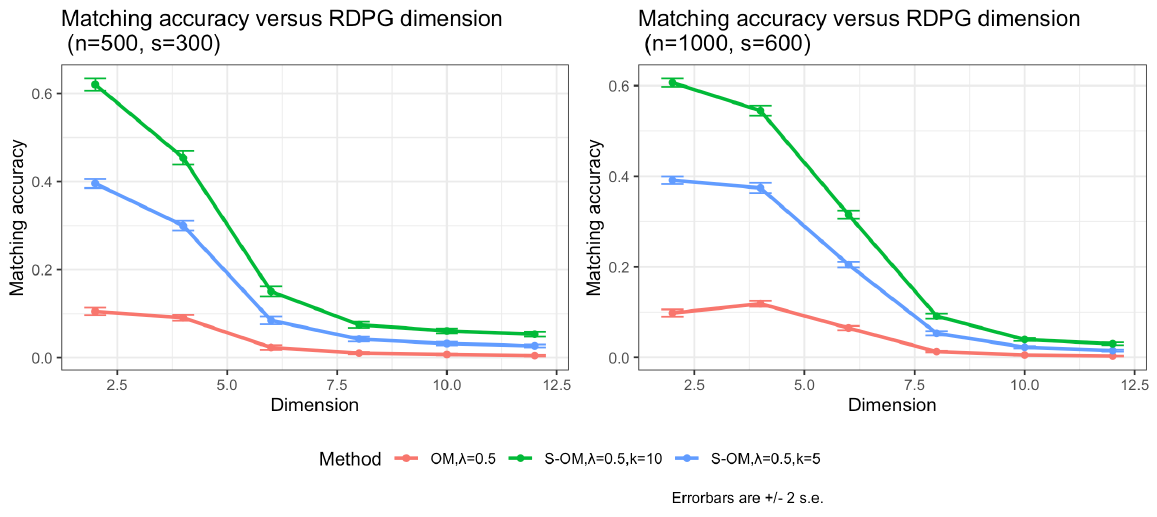}\\
\includegraphics[width=.48\textwidth]{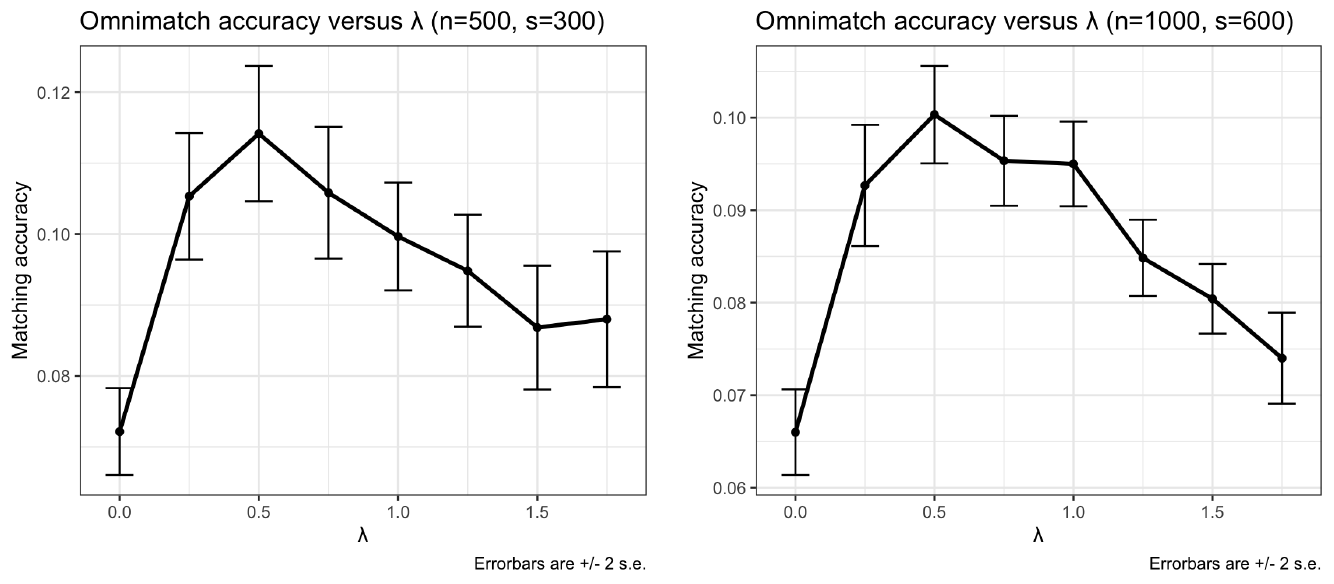}
\caption{Proportion of correctly unshuffled vertices for \texttt{OmniMatch} and \texttt{S-OmniMatch} as a function of model dimension (top) and $\lambda$ (bottom).  Results are averaged over 30 Monte Carlo iterates.
}
\label{fig:dim}
\end{center}
\end{figure}

\subsection{\texttt{OmniMatch} for graph matching}
\label{sec:match}

To evaluate the matching accuracy of the \texttt{OmniMatch} algorithm, we first generate vertex-aligned graph pairs from the $d$-dimensional RDPG, where the i.i.d. latent positions are obtained by projecting a Dirichlet($\vec{1}_{d+1}$) random vector onto the first $d$ coordinates.
We then shuffle a subset of the vertices in one graph (treating the remaining vertices as seeds), apply \texttt{OmniMatch}, \texttt{SGM}, and \texttt{ProcMatch} and compute the proportion of correctly aligned vertices.
The adjacency matrices are generated with $n=s+u=500,\,1000,\,10000$
%,\,200000$
vertices, and we consider varying values of $s$ and $u$ for each $n$.
The proportion of correctly matched vertices after shuffling and embedding is computed for two different variants of \texttt{OmniMatch} and \texttt{ProcMatch}. 
The first method uses the ``hard'' matching approach presented above, where the estimated latent positions of the shuffled vertices are matched using a linear assignment solver. 
In this case, a shuffled vertex is correctly matched after unshuffling if it is exactly matched to its original position. The second method uses a ``soft'' matching approach (call this \texttt{S-OmniMatch}, and \texttt{S-ProcMatch}) where 
a shuffled vertex is correctly matched if its original unshuffled position is among the $k$ nearest neighbors in the embedded space. 
In each case, for each choice of parameters, we average over $nMC=30$ 
simulations.

In Figure \ref{plt:exp_prop_n500} we show results for $n=500,$ and $1000$ vertices.
In both cases, we observe that the accuracy of all algorithms increase with the number of (nested) seeds, with \texttt{SGM} performing worse than hard \oo and \texttt{ProcMatch}.
As noted in Remarks \ref{rem:remlambda}--\ref{rem:remproc}, in both cases we see that \oo with $\lambda=0.5$ outperforms the $\lambda=0$ case, and that \oo with $\lambda=0.5$ achieves nearly identical performance to \texttt{ProcMatch} in the soft and hard matching setting.
In the soft-match setting (implemented in \texttt{SGM} via the soft matching returned by the \texttt{gm()} function in \texttt{iGraphMatch}), we see that \oo and \texttt{ProcMatch} again significantly outperform \texttt{SGM}, with nearly 60\% accuracy achieved in both settings for modest soft-matching level and seed number.
Note that these are matching 2-d RDPGs (and we use the true embedding dimension here in our embedding algorithms); to see the effect of $d$ on matching accuracy, see Figure \ref{fig:dim}.
We see in Figure \ref{fig:dim} that for larger graphs, performance of hard \oo increases with $d$ initially before decreasing for larger $d$.
This coincides with our theory, as larger $d$ allows for a better growth rate of $u$ in terms of $s$, though the constants in our probability bounds often increase in $d$ which eventually will inhibit performance given a fixed $n$ and $s$.
In Figure \ref{fig:dim}, we also see the effect of $\lambda$ on accuracy (here in a 2-d model).  We see that too small or too large a value leads to worse performance, but that performance is fairly stable across small--to--medium $\lambda$ values.
We see a similar phenomena when $n=10000$ where $\lambda=0.5$ outperforms $\lambda=0$; see Appendix A.2.1, Figure 
9--10
% \ref{plt:exp_prop_n1000}--\ref{plt:exp_prop_precision_k} 
for detail.
Notably, in the $n=10000$ case, we see the effect of dimension play out (as predicted in the theory) with increased matching in the $d=10$ case versus the $d=2$ case, and worse accuracy in higher dimensions ($d=15$ here).
\begin{figure}[t!]
\begin{center}
    \includegraphics[width=.48\textwidth]{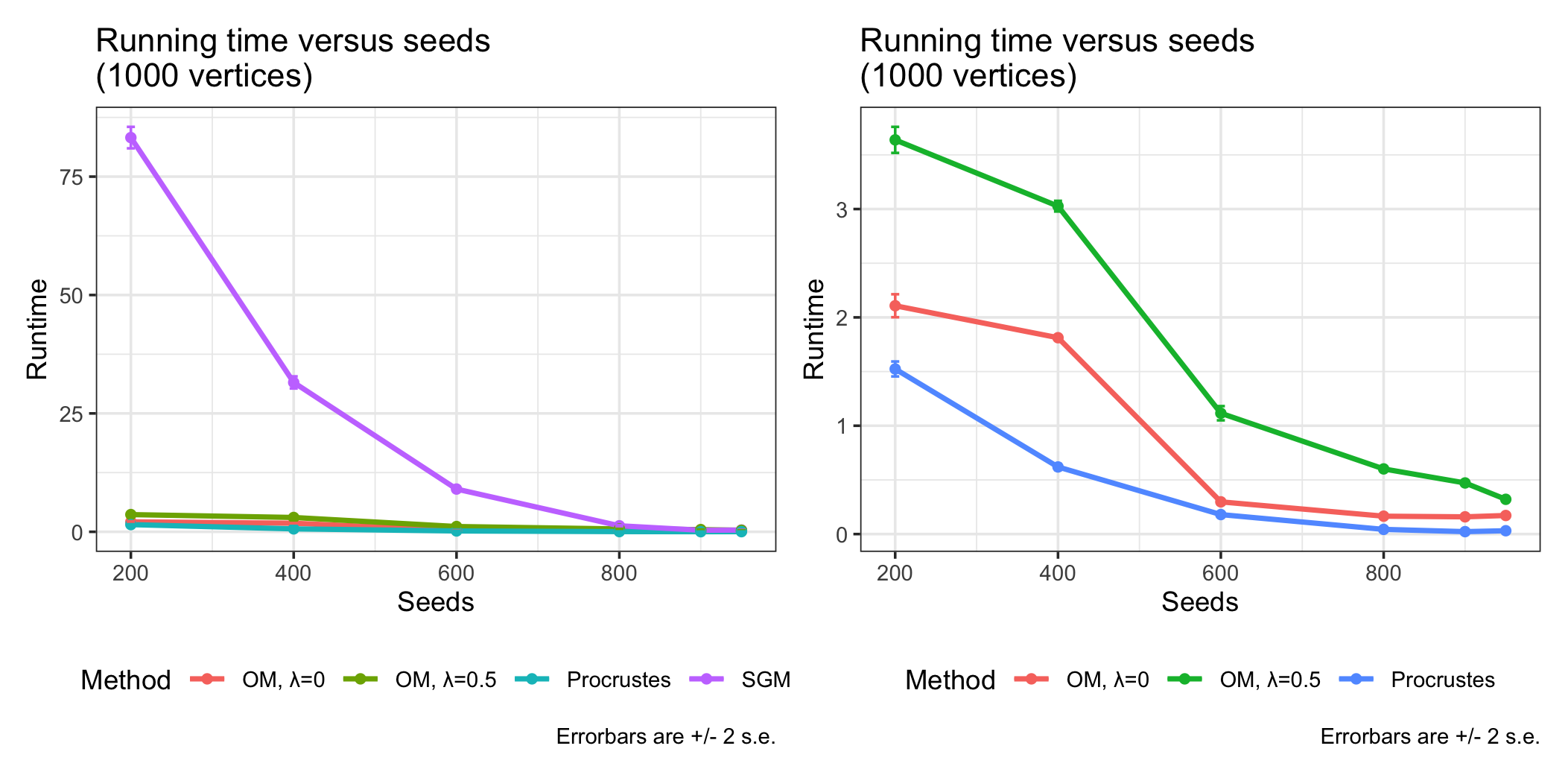}
  \caption{Running times (in sec.) for each algorithm in the $n=1000$ case.
  Results are averaged over 30 Monte Carlo iterates.}
  \label{plt:rt}
\end{center} 
\end{figure}
\begin{rem}
\emph{
\oo (and \texttt{ProcMatch}) require seeds in order to be implemented.
While seeds are not necessary for network alignment procedures \cite{vogelstein:_fast}, even a few seeds have been demonstrated to dramatically increase matching performance \cite{fishkind09:_seeded}.
A natural choice for unseeded spectral graph matching is the classical \texttt{Umeyama} algorithm of \cite{umeyama} (as implemented in \texttt{iGraphMatch}).
With no seeds, we see that in the 1000 vertex setting of Section \ref{sec:match}, the \texttt{Umeyama} algorithm correctly matches on average (averaged over 30 iterates) $\approx 4.8$ of the 1000 vertices.
With soft-matching (here at level $k=9$), \texttt{Umeyama} aligns on average $8.8$ vertices correctly.
With a small number of seeds (here $s=50$), \oo soft-matches (again at level $k=9$) on average $\approx 31.0$ vertices correctly.
As in \cite{fishkind09:_seeded}, we see marked improvement with only relatively few seeds, with performance monotonically increasing as more seeds are used (see Fig. \ref{plt:exp_prop_n500}).
}
\end{rem}

\begin{rem}
    \emph{Running times in the $n=1000$ case for each algorithm are plotted in Figure \ref{plt:rt}.
    In the figure, we see that \texttt{SGM} is markedly the slowest algorithm, with \texttt{ProcMatch} marginally faster than \oo.
    We also see that when $u$ is small (resp., large) relative to $s$, then \oo with $\lambda=0.5$ and with $\lambda=0$ are roughly the same speed (resp., $\lambda=0$ is faster) as predicted by the running time analysis in Section \ref{sec:omnimatch}.}
\end{rem}

\begin{figure}[t!]
\begin{center}
    \includegraphics[width=.48\textwidth]{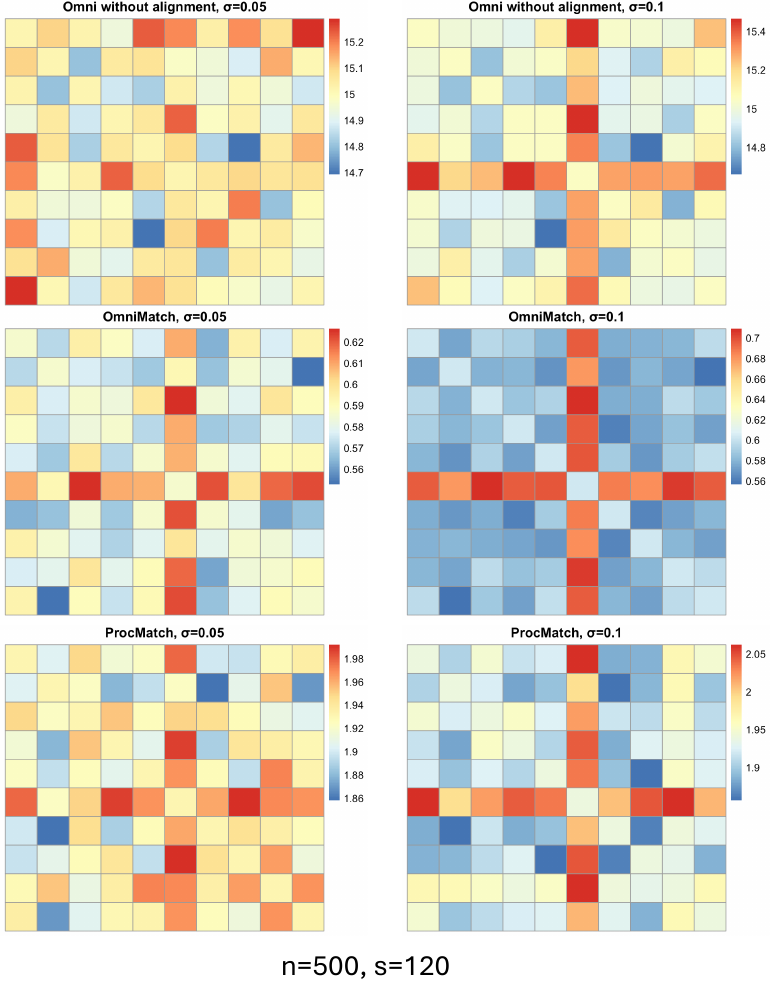}
  \caption{Heat maps of the pairwise distances among 10 RDPGs with anomalous error $\sigma=0.05$ (L) and $0.1$ (R).
  Methods employed are 
  \oo (middle row), \texttt{ProcMatch} (bottom row), and the \texttt{Omnibus} method without label matching (top row).
  Results are averaged over 30 Monte Carlo iterates.
Note that diagonals are imputed as the average of all off-diagonal entries to allow for a more stable color gradient in the heat maps.}
  \label{plt:exp_multi_rdpg}
\end{center} 
\end{figure}

\subsubsection{\texttt{OmniMatch} for multiple graph matching}
\label{sec:multimatch}

Graph matching can be used to compute pairwise graph similarities: given a collection of unaligned (or partially aligned) networks, we can use graph matching tools to align the networks and use the graph matching objective function as a proxy for similarity of the networks.
The matching objective function of Eq. \ref{eq:mof} is a commonly occurring pseudo-distance \cite{bento2018family} in the literature, upper bounds the popular cut metric, and (properly normalized) is the root of the total correlation/alignment strength measure of graph similarity from \cite{fishkind2019alignment}.
To demonstrate the capability of $\oo$ to simultaneously align multiple graphs---and hence simultaneously provide similarity measures across a graph collection---we consider the following illustrative experiment in anomaly detection in a collection of RDPGs. 
We generate 10, 500 node RDPGs from the same distribution, which posits i.i.d. latent positions from the Dirichlet($\vec{1}_{3}$) distribution projected onto the first $d=2$ coordinates.
Each graph contains 500 nodes and we introduce $s \in\{ 60, 120\}$ seeded nodes for every graph, and shuffle the remaining nodes (the shuffling permutations differ graph--to-graph). 
To simulate errors, we add error to graph 6:  We select 80 of its shuffled vertices at random; we then add errors into its latent positions (i.i.d. Gaussian$(0,\sigma)$ noise for $\sigma\in\{0.05, 0.1\}$).
We then embed the collection, and compute pairwise graph distances using \oo (with $\lambda=0.5$), \texttt{ProcMatch}, and the \texttt{Omnibus} method without label matching;
results are visualized in Figure \ref{plt:exp_multi_rdpg} (for the analogous results when $s=60$, see Figure 
11
% \ref{plt:exp_multi_rdpg2} 
in Appendix A.2.2). 
Here, in \oo the seeded vertices are simultaneously embedded across all 10 networks, and \oo used to pairwise align the  unseeded vertices.
\texttt{ProcMatch} is applied pairwise, and from Figure \ref{plt:exp_multi_rdpg} we see that the borrowed strength in the \oo seed embedding leads here to more stable matching results; manifested here in the anomalous graph being better differentiated (i.e., having uniformly higher distance).
We see here that both aligned methods (unsurprisingly) outperform the \texttt{Omnibus} method without label matching, although with enough error even the unaligned method identifies the anomaly.

\subsubsection{Matching and clustering in brain data}
\label{sec:brains!}

We next explore the methods of Section \ref{sec:multimatch} in a connectomic data set. 
The data set consists of 10 DTMRI brain scans for each of 30 subjects from the HNU1 data repository at \href{https://neurodata.io/mri/}{neurodata.io}. These connectomes data are produced with NeuroData's MRI Graphs pipeline, m2g \cite{chung2021low}. Each scan is represented as a weighted graph with 70 vertices after being registered to the Desikan brain atlas. 
Each vertex corresponds to a region of interest (ROI) from the brain atlas and the edges between vertices indicate the strength of neuronal connections between these regions. All the graphs are pre-aligned both within individual subjects and across subjects. 

\begin{figure}[t!]
\begin{center}
  \includegraphics[width=0.48\textwidth]{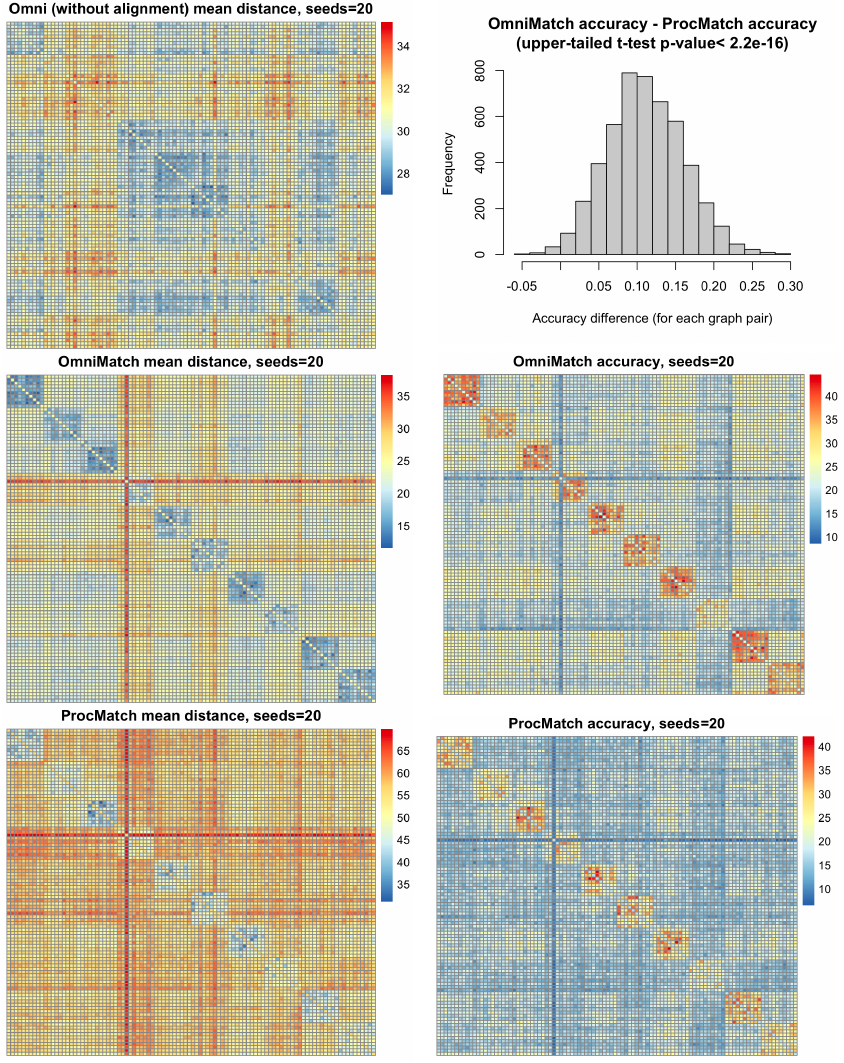}
   
  \caption{Mean pairwise distance (L) and mean matching accuracy (R) heatmaps among 100 brain scans. A total of 10 subjects, each with 10 brain scan-derived graphs, are included for embedding. 
  Methods employed are 
  \oo (middle row), \texttt{ProcMatch} (bottom row), and the \texttt{Omnibus} method without label matching (top row).
  Results are averaged over 30 Monte Carlo iterates.
  The top right panel shows a histogram of pairwise matching accuracies for \oo minus \texttt{ProcMatch}, along with a p-value for the upper-tailed t-test for testing if the mean difference is greater than 0.}
  \label{plt:brains}
\end{center}
\end{figure}

In the experiments, we select the first 10 subjects, resulting in a total of 100 brain scans. Using the automatic dimensionality selection method (locating an appropriate elbow of the SCREE plot) based on \cite{zhu2006automatic,chatterjee2014matrix}, we set the embedding dimension to $dim = 7$ for these brain scan graphs. For this set of graphs, we randomly shuffle
50 vertices across all the graphs. 
We consider three approaches.
As in Section \ref{sec:multimatch}, we compute the pairwise graph distances across each pair in the collection via \texttt{OmniMatch}, \texttt{ProcMatch}, and
the \texttt{Omnibus} method without alignment.
This whole process is repeated for $nMC = 30$ rounds, and the average pairwise distances are presented in the left column of Figure \ref{plt:brains}; note that in the heatmaps, the 100 graphs are ordered by subject, with each subject contributing 10 scans; thus the diagonal blocks correspond to within-subject comparisons. Blue colored blocks along the diagonal indicate that scans from the same subject exhibit shorter pairwise distances, reflecting higher similarity.  
These heatmaps (and the associated Adjusted Rand Indices (ARI's) \cite{rand1971objective} in Table \ref{table:ARI}  comparing the heatmap clustering---using \texttt{hclust}---to the true labels) show one of the potential drawbacks of \oo's borrowed strength: When the graphs are sufficiently different (as is the case with brain scans from different subjects), \oo will force the collection of seeds (and hence the embeddings) to be more similar than they would be otherwise (see \cite{pantazis2022importance} for more on the \texttt{Omnibus} method's induced correlation).
This results in slightly lower clustering performance for \oo as compared to \texttt{ProcMatch}, though the difference is not significant.
That said, \oo is designed for vertex-level alignment (for which the borrowed strength is beneficial), and we see that the accuracy of the \oo vertex recovery is significantly higher than that of \texttt{ProcMatch} across the brain collection (see the first panel in column 2 of Figure \ref{plt:brains}).

\begin{table}[t!]
\caption{Mean ARI for clustering and accuracy for matching brain scans ($nMC = 30$ simulations)}
\label{table:ARI}
\centering
\begin{tabular}{ c |c c c c }

\hline
 Algorithm  & \texttt{Omnibus} & \texttt{OmniMatch} & \texttt{ProcMatch}  \\
 & (w/o align)& & \\
 \hline %\hline

avg(ARI)$\pm$2s.e. &0.026$\pm$0.006 & 0.542$\pm$0.032 & 0.576$\pm$0.046 \\
\hline
avg(Accuracy) &NA & 0.460 & 0.349 \\

\hline
\end{tabular}
\end{table}

\subsection{Power Test With Shuffling}
\label{sec:test}

We next aim to explore the effect of \texttt{OmniMatch} on two-sample graph hypothesis testing.

\subsubsection{Graph hypothesis testing}
\label{sec:hypotest}

Graph hypothesis testing aims to determine whether two samples of graphs are generated from the same underlying distribution.
Researchers have proposed a range of methodologies and inferential techniques for graph hypothesis testing including the work in \cite{ghoshdastidar2020two,ghoshdastidar2018practical, yuan2023practical, tang2017nonparametric,tang14:_semipar, du2023hypothesis,nguen2026network}. 
In the context of RDPG's, a semiparametric two sample graph hypothesis testing framework can be framed as follows.
Given two mutually independent graph samples $(\bA^{(1)},\dots, \bA^{(k_1)} , \bX )\sim \mathrm{JRDPG}(F_X,n)$ and $(\bB^{(1)},\dots,\bB^{(k_2)}, \bY )\sim \mathrm{JRDPG}(F_Y,n)$, we consider the hypothesis testing problem:
\begin{align*}
    H_0: \bX \eer \bY; \quad H_A: \bX \neer \bY 
\end{align*}
where $\bX \eer \bY$ holds if there exist an orthogonal matrix $\bW \in \mathcal{O}_d$ such that $\bX = \bY\bW$.
This accounts for the rotational non-identifiability of the latent positions in the $\mathrm{RDPG}$ model. 
Given estimated latent positions $\hat{\bX}$ and $\hat{\bY}$ (computed, for example, via the ASE), there are a number of natural test statistics that can be formulated; for example, a scaled version of $T = \| \hat{\bX} - \hat{\bY}\bW \|_F$ is used in \cite{levin2017central,tang14:_semipar} and a scaled version of $T = \| \hat{\bX}\hat{\bX}^T - \hat{\bY}\hat{\bY}^T\|_F$ is used in \cite{saxena2025lost}.
In the RDPG framework, suitable bootstrapping \cite{levin2025bootstrapping,tang14:_semipar} can be used to estimate the appropriate critical value and testing power.

Most existing testing methods assume that the nodes across graphs are perfectly aligned, and these methods cannot be readily applied when there are errors across labels. 
However, in many real-world applications the identity of nodes is often unknown, inconsistent, or partially observed, which leads to unaligned or partially aligned graphs.
As discussed in \ref{sec:aligned}, these errorful node correspondences can have a detrimental effect on testing power \cite{saxena2025lost}.
Defining the noisy test in this setting is nuanced.
 Considering the simplified setting where $k_1=k_2=1$, let the potential label error be represented by a fraction, say $\leq k/n$, of the vertices being shuffled.
We then observe $\bA$ and $\widetilde{\bQ} 
 \bB\widetilde{\bQ} 
 ^T$ for some $\widetilde{\bQ}\in\Pi_{n,k}=\{\bP\in\Pi_n:\text{tr}(\bP)\geq n-k\}$.
For each $\bQ\in\Pi_{n,k}$, the associated testing critical value is defined as the smallest value $c_{\alpha,\bQ}$ such that (where $T$ is the appropriate test statistic)
$\mathbb{P}_{H_0}(T \geq c_{\alpha,\bQ})\leq \alpha.$
As the observed $\widetilde{\bQ}$ is unknown, the conservative shuffled test would reject the null if $T \geq \max_{\bQ\in\Pi_{n,k}}c_{\alpha,\bQ}$; this can result in lost testing power if $c_{\alpha,\widetilde{\bQ}}<\max_{\bQ\in\Pi_{n,k}}c_{\alpha,\bQ}$ which can occur, for example, if $\widetilde{\bQ}$ shuffles less than $k/n$ fraction of the vertices \cite{saxena2025lost}.
When node alignments are not observed or properly recovered via graph matching, the underlying structural similarity between graphs may be obscured, and traditional testing methods can experience a significant loss of statistical power or inflated false positive rates.

\begin{figure}[t!]
\begin{center}
\vspace{-5mm}
  %\subfloat[err$ = 0.01$]
  \includegraphics[width=0.48\textwidth]{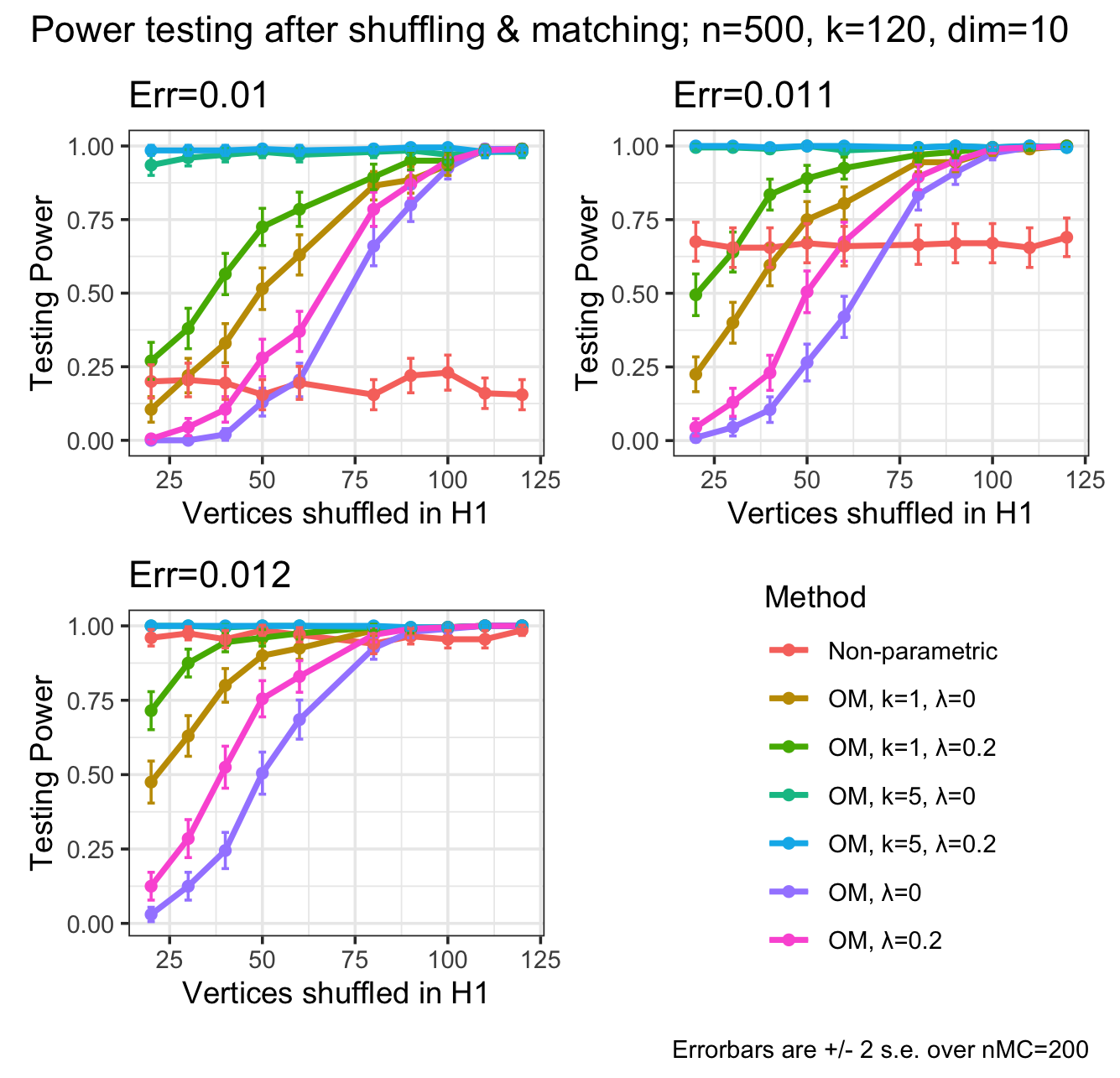}
\caption{
Statistical power for two-sample graph hypothesis testing under node shuffling. 
Panels display results from experiments on pairs of graphs with $n=500$ nodes. In each, one graph is fixed while the other has noise $\textrm{err} \in {0.01,0.011,0.012}$ added to the latent positions. All embeddings have dimension $d=10$. 
Each panel
compares the empirical testing power of \texttt{OmniMatch}, \texttt{S-OmniMatch} using 1 nearest neighbor ($k=1$), using 5 nearest neighbors ($k=5$), and a non-parametric test. }
\label{plt:exp_rdpg_d10}
\end{center}
\end{figure}

\subsubsection{OmniMatch as a preprocessing tool}
\label{sec:omnitestresults}

In the shuffled testing setting, one option is to use \texttt{OmniMatch} (and \texttt{S-OmniMatch}) as a preprocessing tool before applying more classical testing methods that assume known alignment \textit{a priori.}
Another option is to ignore the signal/noise in the labels across graphs and use a nonparametric (i.e., label free) testing procedure such as that proposed in \cite{tang14:_nonpar,alyakin2024correcting}.
Similar to Section \ref{sec:match}, we consider $n=500$ vertex RDPG's, where under $H_0$ the graphs are generated with i.i.d. 
Dirichlet($\vec{1}_{d+2}$) 
latent positions projected onto the first $d$ coordinates.
 With notation as in Section \ref{sec:hypotest} (here $k_1=k_2=1$), under the null hypothesis, $\bA$ is observed cleanly while  $v_0 = 120$ vertices of $\bB$ are shuffled. Under the alternative hypothesis, $v_1 \in \{20, 30, 40, \ldots, 100, 110, 120\}$ vertices are shuffled in $\bB$. At the same time, a different scale of error is added to the latent positions $\bY=\bX+\mathrm{err}$ as noise, where $\mathrm{err} = 0.01, 0.011, 0.012$. 

 We employ two methods for the parametric tests. Graph pairs are generated with and without shuffled vertices corresponding the hard-matched \texttt{OmniMatch} and the soft-matching \texttt{S-OmniMatch} with $k=5$; in the soft-matched case, the associated estimate for the soft-matched latent position is the average of these $5$ soft-matched vertices.
For both methods, the Frobenius norm between the latent position estimations of the whole graph pairs is computed as the test statistic (as in \cite{levin2017central,tang14:_semipar}). 
We conduct  $nMC = 200$  rounds of simulation under the null and determined the $95^{th}$  percentile of the null hypothesis distribution as the critical value. The power of the test is computed as the proportion of distances exceeding the critical value over the $200$ simulations under the alternative model. 
We compare the performance of these two methods with the non-parametric test proposed in \cite{tang14:_nonpar,alyakin2024correcting}, which is applied to pairs of graphs directly without need for pre-matching (this method estimates and compares the latent position distributions and do not use the labels across graphs). 
All the embedding methods and tests are incorporated using \textbf{graspologic} Python package.

Results for $d=10$ are displayed in Figure \ref{plt:exp_rdpg_d10}; note that similar results hold when $d=2$ (see Figure 
13
% \ref{plt:exp_rdpg_d2} 
in Appendix A.3) though this 
higher embedding dimension setting allows for more structural information about the graphs to be preserved in the embedding space, leading to improved discriminative power (in these settings, for computational efficiency we used the following parameters in the gradient descent for \oo: learning rate = 1e-3, tol = 1e-5, max$\_$iter = 500).
Note also that here $\lambda=0.2$ provides the highest testing power for \oo---see Figure 
12 
% \ref{plt:exp_rdpg} 
in Appendix A.3
for a comparison of power across $\lambda$---and in Figures \ref{plt:exp_rdpg_d10} and 13
% \ref{plt:exp_rdpg_d2} 
we include only the $\lambda=0.2$ case to unclutter the figures.
We see that, in general, \texttt{S-OmniMatch} outperforms \texttt{OmniMatch} across most parameter settings.
Moreover, soft matching with 5 nearest neighbors consistently achieves higher power compared to using 1 nearest neighbor; using 5 nearest neighbors achieves nearly optimal testing power across all parameter settings.
As expected, the nonparametric method exhibits increasingly good power as $\textrm{err}$ increases.
What is of interest is that in these higher power settings, the matching-based tests (\texttt{S-OmniMatch} with $k=1$ and \texttt{OmniMatch}) are worse than the nonparametric tests in the overly conservative testing regime ($v_1<v_0$), though the testing power increases and surpasses the nonparametric alternative as $v_1$ increases to $v_0$.
Exploring this intersection point is of great interest, and is the subject of ongoing work.

\subsection{OmniMatch applied to English-Zulu Parallel Sentences}
\label{sec:engzulu}

We further conduct experiments on a cross-lingual dataset consisting of 1000 English sentences and their corresponding translations into Zulu.
Sentence embeddings were obtained using the Nomic embedding framework \cite{nussbaum2024nomic}, applied to the raw textual data.  
The resulting datasets for both English and Zulu consist of matrices of size $1000 \times 768$, where each row corresponds to the 768 dimensional embedding of a sentence. For each language, we compute the pairwise cosine similarity matrices within each data set and use them as input for the \texttt{SGM}, $\oo$ and \texttt{ProcMatch} algorithms. 
The use of cosine between sentence or document vectors is well established within Information Retrieval, and has found robust practical application since the late 1970s \cite{wong1984vector}.
While these graphs are not RDPGs, we suspect that the similarity structure here is amenable to the same theoretical spectral analysis as in the RDPG setting; see \cite{levin2026adjacency} for similar embedding analysis of correlation networks.  Moreover, these results demonstrate the effectiveness of \oo in broader network settings. 
This results in $m=2$ graphs whose vertices/nodes are $n$=1000 sentences, each of which has a corresponding translation in the other graph. The edges are inferred sentence-level measures of semantic similarity computed in the embedding spaces.
Following a similar experimental setup as described in Section \ref{sec:match}, we evaluate matching performance under varying levels of node shuffling, with the number of seeds set to $u = [950, 900, 800, 600, 500, 400]$ embedding dimensions chosen from $dim = [2,10,50]$. 
Here a correct match corresponds to recovering the correct unseeded Zulu translation for each unseeded English sentence.
Results are averaged over $nMC = 30$ simulations (yielding $30$ random, nested seed sets), and the average matching accuracies are computed and visualized in Fig. \ref{plt:engzulu}. 
Note here that $\lambda=0.5$ appears to be the best choice for matching accuracy (see Fig. 
14
% \ref{fig:lls} 
in Appendix 
A.4
% \ref{sec:extraZulu} 
for detail), and we omit the other $\lambda$ values in this section to help de-clutter the figures.

In previous examples, we used an automated dimension selection routine to choose our embedding dimension $d$ based on locating an appropriate elbow in the SCREE plot.
\begin{figure}[t!]
\begin{center}
    \includegraphics[width=.47\textwidth]{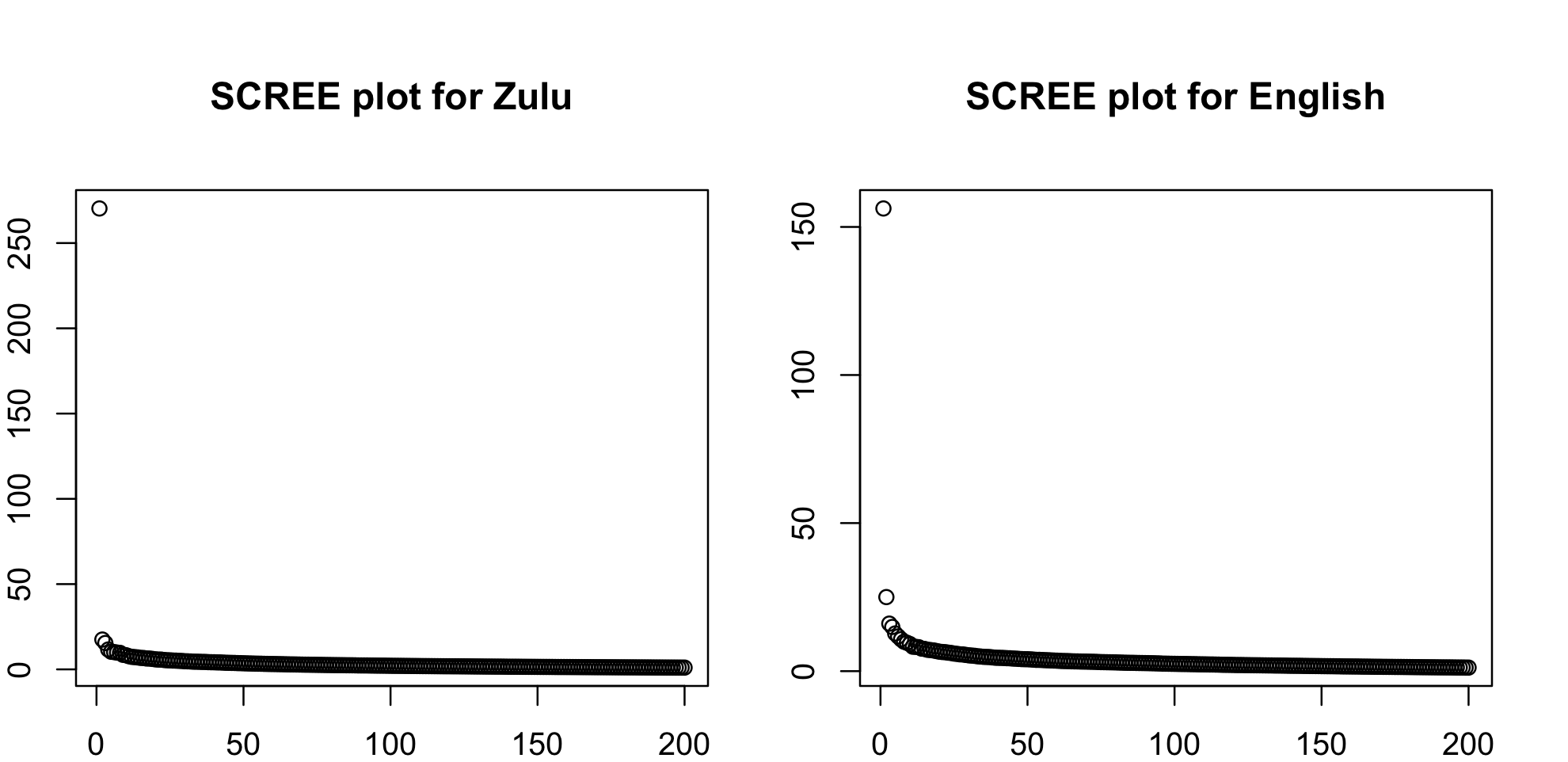} 
\caption{
SCREE plots for the Zulu and English similarity matrices.}
\label{fig:SCREE}
\end{center}
\end{figure}
In Figure \ref{fig:SCREE}, we see that each matrix has a single dominant singular value, and our automated procedure finds a second elbow at the 25th (resp., 24th) singular value for the Zulu (resp., English) matrix.
In Fig. \ref{plt:engzulu}, we plot the average graph matching accuracies using \texttt{SGM}, $\oo$ and \texttt{ProcMatch} (hard matching and with 10-nearest neighbor matching). 
While the dominant singular value in the SCREE plot may lead us to think that a lower embedding dimension is preferable, this is not the case here.  Indeed, we see performance of 
$\oo$ and \texttt{ProcMatch} increase dramatically from $d=2$ to $d=10$ to $d=25$, and a modest increase from $d=25$ to $d=50$ and from $d=50$ to $d=100$ (at least for \texttt{ProcMatch}, though \oo does see a slight performance increase as well in the higher dimensions).  While \texttt{SGM} uses the full similarity matrix, we see that accuracy here can be increased by matching in a lower dimensional space.
Here, when $d=100$, $\oo$ and \texttt{ProcMatch} perform nearly identically, both outperforming \texttt{SGM}.
Notably, these graphs are particularly hard to align, as the difference in objective function value between the true permutation and a random permutation is relatively quite small (see Figure 
15
% \ref{fig:hists} 
in Appendix 
A.4
% \ref{sec:extraZulu}
).
While the original data here is $768$ dimensional, we see here that performance increases for all $d$ considered, though $d=50$ and $d=100$ show similar performance for \oo.
This suggests that the alignment signal in the data is captured in higher dimensional structure than would normally be considered in our SCREE plot analysis, though, as expected, there appears to be a diminishing return when including many data dimensions.

\begin{figure}[t!]
\begin{center}
    %\subfloat[err$ = 0.012$]
    \includegraphics[width=.47\textwidth]{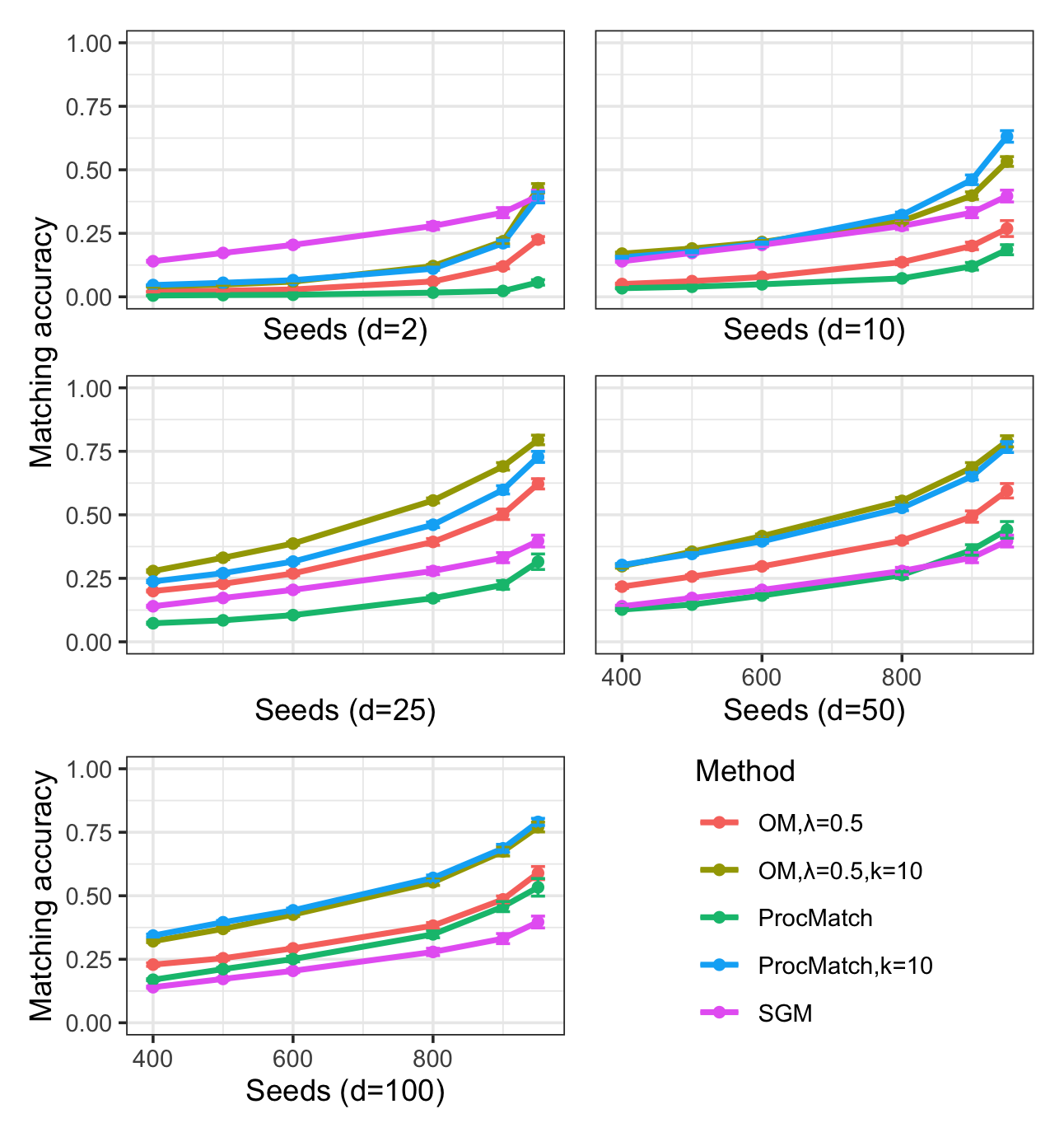} 
\caption{
Graph matching accuracies for \texttt{SGM}, \texttt{ProcMatch} (hard and soft) and \texttt{OmniMatch} (hard and soft) on English--Zulu data. The top (resp., middle, bottom) panel presents the mean matching accuracies using embedding dimension 2 (resp., 10, and 50) for \texttt{ProcMatch} and \texttt{OmniMatch}.
Results are averaged over 30 Monte Carlo replicates and error bars are $\pm2$s.e.}
\label{plt:engzulu}
\end{center}
\end{figure}

\section{Discussion}
\label{sec:discuss}

In this work, we propose \texttt{OmniMatch} to address the challenge of partial node misalignment in graphs.
Beyond proposing an algorithm for efficient graph node matching, we also consider the impact that label uncertainty and subsequent matching have in the framework of two-sample hypothesis testing. 
Two-sample network testing is just one example of existing graph inference methods that often rely on the assumption of known node correspondences, thereby limiting their applicability when alignment is unknown. To overcome this limitation, \texttt{OmniMatch} integrates joint embedding and approximate graph matching under the setting of independent edge correlations, enabling both node correspondence recovery and robust hypothesis testing across graph pairs.
Leveraging the RDPG model and the Omnibus embedding framework, \texttt{OmniMatch} extends latent positions to out-of-sample nodes and formulates an approximate graph matching problem that can be efficiently solved. The proposed method generalizes beyond pairwise alignment to accommodate the joint matching of multiple ($m>2$) graphs, enabling simultaneous alignment across networks. We theoretically establish the consistency of \texttt{OmniMatch} under general conditions, ensuring its robustness and reliability for a wide range of graph inference tasks.

Through comprehensive simulation results and real-world data analysis, we demonstrated \texttt{OmniMatch} effectively recovers node alignments, improves the power of two-sample tests under varying levels of node shuffling and embedding dimensions, and practically distinguishes the graph with disorganized nodes among a collection of graphs. 
Moreover, it has practical application to problems of harvesting NLP data, such as scraping and aligning parallel sentence pairs from web documents which contain similar content in different languages \cite{banon-etal-2020-paracrawl}.
The results underline the necessity of integrating graph matching with joint embedding on statistical testing when addressing node misalignment. Extending the current framework to fully unaligned structure, dynamic temporal networks, or graphs with varying sizes still desires further investigation. Additionally, optimizing the embedding dimension selection and improving computational scalability for very large graphs applications are also critical for broader applications.

\bibliographystyle{IEEEtran}

\bibliography{biblio,bibliography}


%\begin{IEEEbiography}{}
%
%\end{IEEEbiography}

%\begin{IEEEbiographynophoto}{}

\begin{IEEEbiography}
[{\includegraphics[width=1in,keepaspectratio]{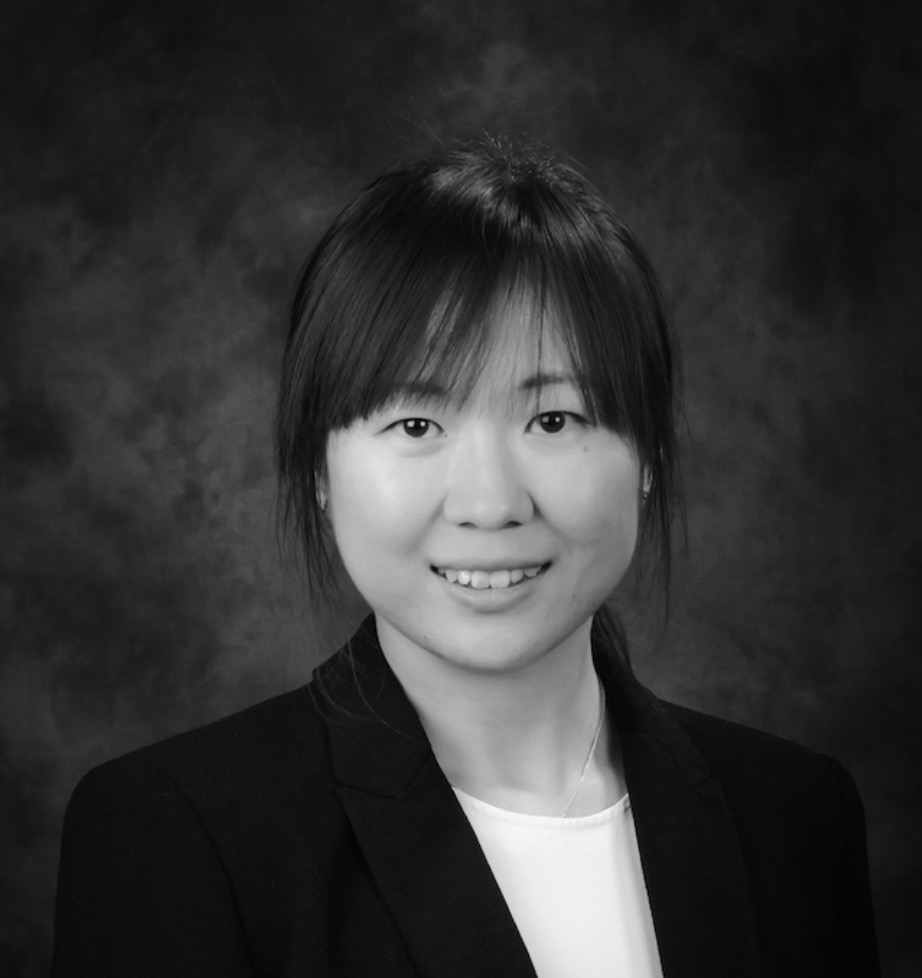}}]{Tong Qi}
received the BS degrees in agricultural economics from Purdue University in 2013, the MS degree in management from Purdue University in 2015, the MS degree in joint statistics and computer science from Purdue University in 2019, and the Ph.D degree from the STAT program at the University of Maryland, College Park in 2025. She is currently a Data Scientist with Capital One, where she works on artificial intelligence and machine learning. Her research interests include graph embedding, statistical network analysis, and machine learning. Webpage: \url{https://tong-qii.github.io/}..
\end{IEEEbiography}

\begin{IEEEbiography}
[{\includegraphics[width=1in,keepaspectratio]{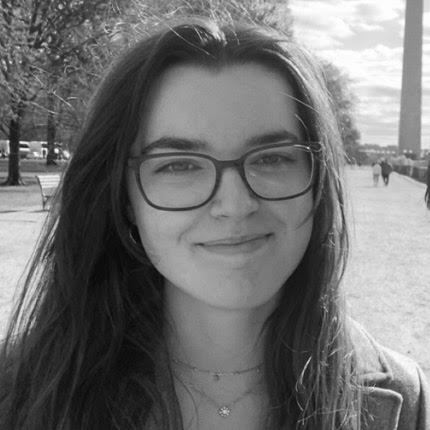}}]{Vera Andersson}
received the BSc and MSc degrees in mathematical statistics from Stockholm University, Sweden, in 2021 and 2023, respectively, and the MA degree in mathematical statistics from the University of Maryland, College Park (UMD), in 2025. Since 2023, she has been pursuing the PhD degree in mathematical statistics at UMD, where she advanced to candidacy in 2025. 
Her research interests include statistical machine learning, statistical inference on random graphs, spectral graph theory, stochastic block models, and geometric network alignment.

\end{IEEEbiography}

\begin{IEEEbiography}
[{\includegraphics[width=1in,height=1.25in,clip,keepaspectratio]{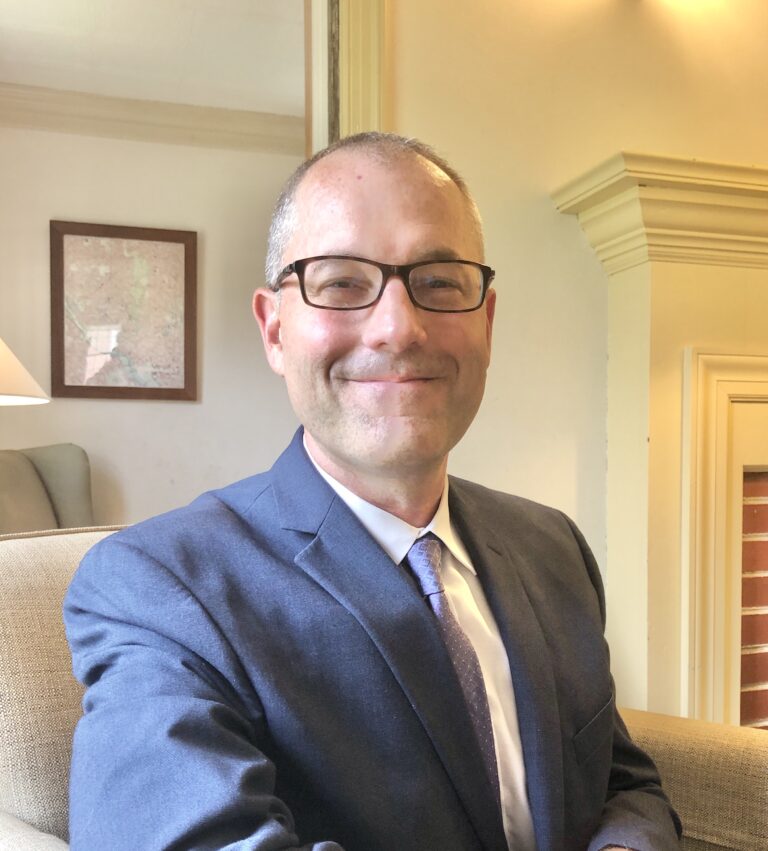}}]{Peter Viechnicki} received the BA degree in Classics and Linguistics from Princeton University in 1992, and the PhD in Linguistics from the University of Chicago in 2002.  After a career
in government and the management consulting industry he was recruited to direct the Human
Language Technology Center of Excellence at the Whiting School of Engineering at Johns Hopkins University from 2020 to 2025. He now works on High Performance Computing. His research interests include computational linguistics, phonetics, and cognitive science.  
\end{IEEEbiography}

\begin{IEEEbiography}[{\includegraphics[width=1in,height=1.25in,clip,keepaspectratio]{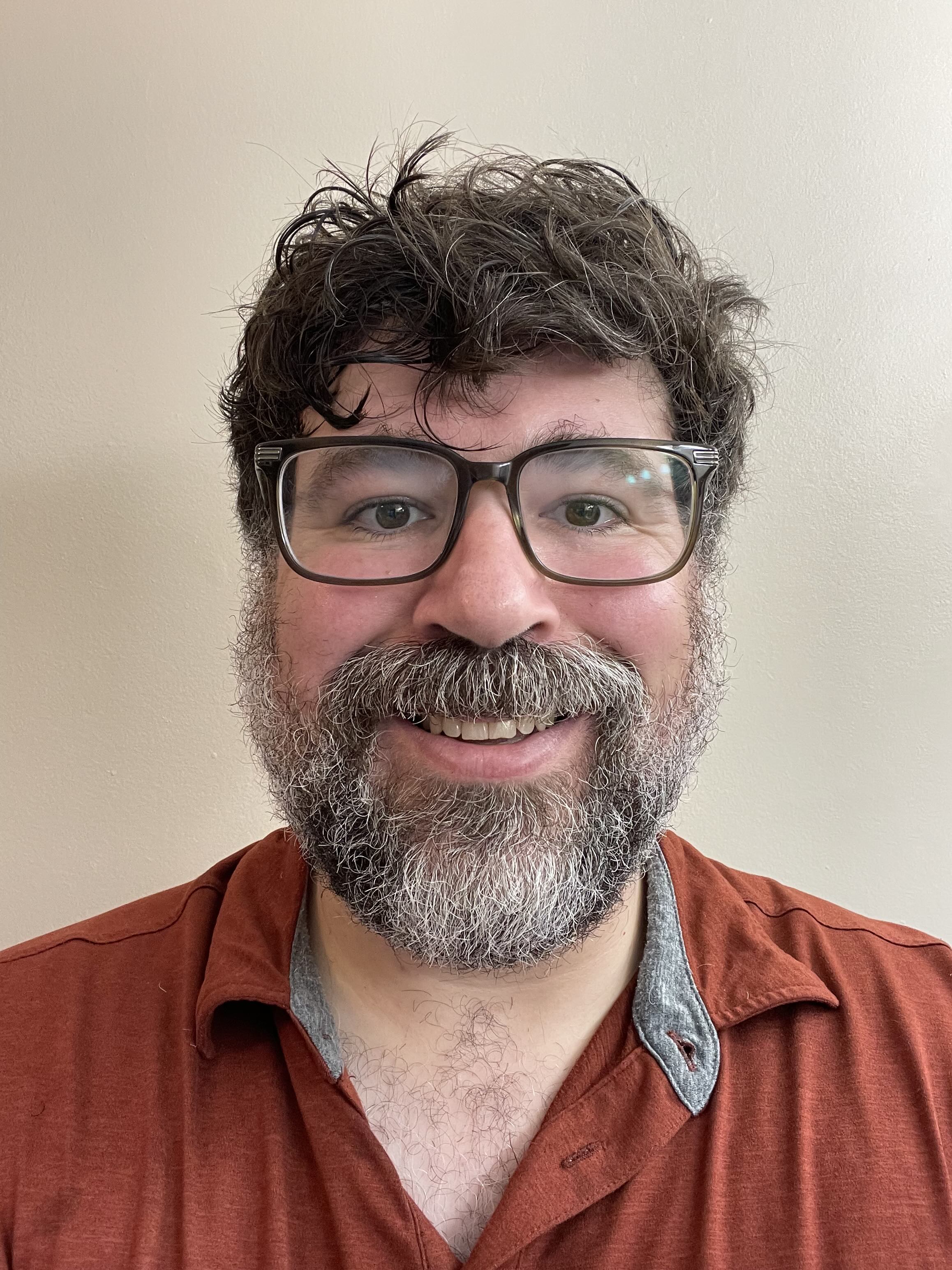}}]{Vince Lyzinski}
 received the BSc degree in mathematics from the University of Notre Dame, in
2006, the MA degree in mathematics from Johns
Hopkins University (JHU), in 2007, the MSE
degree in applied mathematics and statistics
from JHU, in 2010, and the PhD degree in applied
mathematics and statistics from JHU, in 2013.
From 2013-2014 he was a postdoctoral fellow
with the Applied Mathematics and Statistics
(AMS) Department, JHU. During 2014-2017, he
was a senior research scientist with the JHU
HLTCOE and an assistant research professor with the AMS Department, JHU.
From 2017-2019, he was on the Faculty in the Department of Mathematics
and Statistics at the University of Massachusetts Amherst. Since 2019 he has been on the Faculty in the Department of Mathematics at the University of Maryland, College Park, where he is currently a Professor. His research interests include graph matching, statistical
inference on random graphs, pattern recognition, dimensionality reduction, stochastic processes, and high-dimensional data analysis.
\end{IEEEbiography}

\newpage
\onecolumn

\appendix
\renewcommand{\thesubsection}{\Alph{subsection}}

\subsection{Proofs of main results}
\label{app:pf}
Herein we collect the proofs of the main results in this work.
Below, we shall use the following notation.  For a matrix $\mathbf{M}\in\mathbb{R}^{n\times m}$ ($n>m$), we denote the ordered singular values of
 $\mathbf{M}$ via
$$\sigma_1(\mathbf{M})\geq
\sigma_2(\mathbf{M})\geq\cdots\geq
\sigma_m(\mathbf{M}).$$
For a square matrix $\mathbf{M}$ with real spectrum, we will similarly use $(\lambda_i(\mathbf{M}))_{i=1}^n$ to denote the ordered spectrum of $\mathbf{M}$. 

Throughout, we will make use of the following lemmas
\begin{lemma}[Observation 2 in \cite{levin2017central}]
Let $F$ be a distribution on $\mathbb{R}^d$ satisfying the $\delta$-inner product property, and let the $n$-rows of $\Xt$ be i.i.d. $F$ and let $\bY$ independently distributed according to $F$.
With probability at least $1-d^2/n^2$, it holds that for all $i\in[d]$, 
$$|\lambda_i(\Xt\Xt^T)-n\lambda_i(\mathbb{E}(\bY\bY^T))|\leq 2d\sqrt{n\log n}.$$
\label{lem:sigmaX}
\end{lemma}
\noindent Lemma \ref{lem:sigmaX}, combined with Assumption [ii]. on $F$ then provides that there exist constants $C_1<C_2$ (only depending on $F$) such that for $n$ sufficiently large (i.e., for all $n\geq n_0$ some $n_0$ depending only on $F$)
\begin{align}
\label{eq:evalX}
C_1\sqrt{n}\leq \sigma_d(\Xt)\leq \sigma_1(\Xt)\leq C_2\sqrt{n}
\end{align}
holds with probability at least $1-d^2/n^2$.

\vspace{2mm}
\noindent \textbf{Proof road map:} 
The proof of matching consistency will proceed as follows. 
We first establish finite $n$, 2--to--$\infty$ norm  consistency results for the OOS embedding step in \oo.   
This will allow us to show that each OOS point is sufficiently closer to its true latent position than to any other latent position or non-paired (by the latent true alignment) OOS data point.
This will be enough to ensure that---as the latent positions are sufficiently well-separated by assumption---the pairwise MLAP of the OOS points will match each OOS point correctly across graphs.  Details for this are provided below.

\subsubsection{Proof of Lemma \ref{lem:2infOOS}}
\label{app:lem1}

As in \cite{levin2021limit}, we begin with (a slight extension of) Theorem 5.3.1 in \cite{golub2013matrix}.
We state the adapted version here (with our modification) as in \cite{levin2021limit}.

Note that the error term in Eq. \ref{eq:key} is presented in non-asymptotic form here (which requires the slight proof modification provided below).
\begin{theorem}
\label{thm:VL}
Let $\bW$ be as in Lemma \ref{lem:2toinf}.
Suppose that $\ww$, $\wwh\in\mathbb{R}^d$ and $\rr,\rrh\in\mathbb{R}^s$ satisfy
\begin{align*}
\|\Xt\bW \ww-\bb\|&=\min_w \|\Xt\bW w-\bb\|,\quad \rr=\bb-\Xt\bW\ww\\
\|\Xhi \wwh-\bb\|&=\min_w \|\Xhi w-\bb\|,\quad \rrh=\bb-\Xhi\wwh.
\end{align*}
There exist constants $D_0,D_1>0$ and $s_0\in\mathbb{Z}^+$ (all depending only on $F$ and $m$) such that for all $s\geq s_0$, we have that the following holds with probability at least $1-D_0/s^2$:
\begin{itemize} 
\item[i.] We have that 
\begin{align}
\label{eq:2lam}
\|\Xhi-\Xt \bW\|\leq \sigma_d(\Xt)
\end{align}
\item[ii.] $\bb$, $\ww$ are all non-zero (note that to apply Eq. 5.3.11 from \cite{golub2013matrix}, we need not have $\rr\neq 0$ as assumed in \cite{levin2021limit});
\item[iii.] $\Xt\bW$ is rank $d$; 
\item[iv.] Define $\ttt\in(0,\pi/2)$ via $\sin(\ttt)=\|\rr\|/\|\bb\|$.
Letting
$$
\nnn=\frac{\|\Xt\bW\ww\|}{\sigma_d(\Xt\bW)\|\ww\|},
$$
we have (where $\kappa_2(\cdot)$ is the usual condition number of a matrix)
\begin{align}
&\frac{\|\wwh-\ww\|}{\|\ww\|}\notag\\
&\,\,\,\,\leq \frac{\|\Xhi-\Xt\bW\|}{\|\Xt\bW\|}
(1+\nnn\tan\ttt)\kappa_2(\Xt\bW) +D_1\frac{dm\log(mn)^2}{n}\frac{\|\Xhi-\Xt \bW\|^2}{\|\Xt\bW\|^2\|\ww\|}
\label{eq:key}
\end{align}
\end{itemize}
\end{theorem}
We adapt the proof given in \cite{golub2013matrix} here to provide the non-asymptotic form for the error.  This will be key to our subsequent application of the OOS result to multiple vertices simultaneously.
\begin{proof}
Condition on the events in Eq. \ref{eq:2inf} and Eq. \ref{eq:evalX}, so that parts [i]. and [iii]. in the theorem holds immediately.
Indeed, conditioning on the aforementioned events (where $C>0$ is the constant appearing in Lemma \ref{lem:2toinf})
$$
\|\Xhi-\Xt \bW\|^2\leq \|\Xhi-\Xt \bW\|_F^2\leq \sum_{j=1}^s \|(\Xhi-\Xt \bW)_j\|^2\leq C^2 m\log(ms)^2=o(\sqrt{s}).
$$
For part [ii]., by the $\delta$-inner product property assumption, the probability $\bb$ is zero is easily bounded above via Hoeffding's inequality 
by an exponentially small probability
for $s$ sufficiently large (with threshold on $s$ only depending on $F$); indeed we have that if $Y\sim\text{Binom}(s,\delta)$, then
\begin{align*}
    \mathbb{P}\left(\|\bb\|^2=0 \right)\leq \mathbb{P}(Y=0)\leq e^{-2s\delta^2}.
\end{align*}
From $\bb$ being nonzero, we have that $\ww$ would be non-zero as well giving part [ii]. in the theorem.

For part [iv]., we will follow the proof of Eq. 5.3.11 in \cite{golub2013matrix} closely.
Let $$\epsilon=\frac{\|\Xhi-\Xt\bW\|}{\|\Xt\bW\|}$$ and define $$E=\frac{\Xhi-\Xt\bW}{\epsilon}.$$
Theorem 2.5.2 in \cite{golub2013matrix} provides that $\text{rank}(\Xt\bW+t\bE)=d$ for all $t\in[0,\epsilon]$.
The solution $x(t)$ to 
\begin{align}
\label{eq:x(t)}
(\Xt\bW+t\bE)^T(\Xt\bW+t\bE)x(t)=(\Xt\bW+t\bE)^T\bb
\end{align}
is continuously differentiable for all $t\in[0,\epsilon]$, and note that $x(0)=\ww$, $x(\epsilon)=\wwh$.
Differentiating Eq. \ref{eq:x(t)} above, we arrive at the following
\begin{align}
\label{eq:0deriv}
x(t)=\left((\Xt\bW+t\bE)^T(\Xt\bW+t\bE)\right)^{-1}(\Xt\bW+t\bE)^T\bb\\
\label{eq:1deriv}
\dot{x}(t)=\left((\Xt\bW+t\bE)^T(\Xt\bW+t\bE)\right)^{-1}\left(\bE^T\bb-((\Xt\bW)^T\bE+\bE^T\Xt\bW+2t\bE^T\bE)x(t)\right)\\
\label{eq:2deriv}
\ddot{x}(t)=\left((\Xt\bW+t\bE)^T(\Xt\bW+t\bE)\right)^{-1}\left(-2\bE^T\bE x(t)-2((\Xt\bW)^T\bE+\bE^T\Xt\bW+2t\bE^T\bE)\dot{x}(t)\right)
\end{align}
Note next that
\begin{align*}
\left\|\left((\Xt\bW+t\bE)^T(\Xt\bW+t\bE)\right)^{-1}\right\|&=(\sigma_d(\Xt\bW+t\bE))^{-2}\\
&\leq (\sigma_d(\Xt)-t\sigma_1(\bE))^{-2}\\
&\leq\left(C_1\sqrt{s}-\frac{t}{\epsilon}\|\Xhi-\Xt\bW\|\right)^{-2}\\
&\leq
\left(C_1\sqrt{s}-Cm^{1/2}\log(ms)\right)^{-2}
\end{align*}
where in the last line we applied Eq. \ref{eq:2inf}.
For $s$ sufficiently large (bigger than a threshold $s_0'$ depending on $F$ and $m$), we then have that there exists a constant $C_3$ (depending on only $F$ and $m$) such that
\begin{align}
\label{eq:inv}
\|[(\Xt\bW+t\bE)^T(\Xt\bW+t\bE)]^{-1}\|\leq C_3/s
\end{align}
Applying this to Eq. \ref{eq:0deriv}--\ref{eq:2deriv}, we have that for $s$ sufficiently large (bigger than a threshold $s_0''$ depending on $F$ and $m$) there exist constants $C_4,C_5,C_6>0$ (independent of $\bb$) such that for any $t$, 
\begin{align*}
\|x(t)\|
% &= \|[(\Xt+t\bE)^T(\Xt+t\bE)]^{-1}\|\cdot(\|\Xt\|+\|t\bE\|)\cdot\|\bb\|\\
&\leq C_3/s\cdot(C_2\sqrt{s}+Cm^{1/2}\log(ms))\sqrt{s}\leq C_4\\
\|\dot{x}(t)\|&\leq C_3/s\cdot \left(
% \bE^T\bb
C(ms)^{1/2}\log(ms)+
%-x(t)(\Xt^T\bE+\bE^T\Xt+2t\bE^T\bE)
2C_4C_2C(ms)^{1/2}\log(ms)+2C_4C^2m\log(ms)^2\right)\\
&\leq C_5\frac{m^{1/2}\log(ms)}{\sqrt{s}}\\
\|\ddot{x}(t)\|&=
C_3/s\cdot
\left(2C_4C^2m\log(ms)^2+
4C_5C_2C m\log(ms)^2
+2C_5C^2\frac{m^{3/2}\log(ms)^3}{\sqrt{s}}\right)\\
&\leq C_6\frac{m\log(ms)^2}{s}
\end{align*}
A Taylor expansion then yields
$$\wwh=\ww+\epsilon\dot{x}(0)+\epsilon^2R_2$$
where the remainder term $R_2$ satisfies
$$\|R_2\|\leq C_6\frac{dm\log(ms)^2}{2s}$$ from the above bound on $\|\ddot{x}(t)\|$ (which we note is uniform on $t\in[0,\epsilon]$).

This yields that 
\begin{align}
\label{eq:bnd}
    \frac{\|\wwh-\ww\|}{\|\ww\|}=\epsilon \frac{\|\dot{x}(0)\|}{\|\ww\|}+\epsilon^2 \frac{\|R_2\|}{\|\ww\|}.
\end{align}
Now
\begin{align*}
\dot{x}(0)&=
\left((\Xt\bW)^T(\Xt\bW)\right)^{-1}\left(\bE^T\bb-((\Xt\bW)^T\bE+\bE^T\Xt\bW)\ww\right)\\
&=\left((\Xt\bW)^T(\Xt\bW)\right)^{-1}\bE^T\rr -\left((\Xt\bW)^T(\Xt\bW)\right)^{-1}(\Xt\bW)^T\bE\ww
\end{align*}
We then have that
\begin{align*}
\|\dot{x}(0)\|&\leq 
\left\|\left((\Xt\bW)^T(\Xt\bW)\right)^{-1}\bE^T\rr\right\| +\left\|\left((\Xt\bW)^T(\Xt\bW)\right)^{-1}(\Xt\bW)^T\bE\ww\right\|\\
&\leq \frac{\|\Xt\|\cdot\|\rr\|}{\sigma_d(\Xt)^2}+\frac{\|\Xt\|\cdot\|\ww\|}{\sigma_d(\Xt)}
\end{align*}
Plugging this into Eq. \ref{eq:bnd} yields that
\begin{align*}
\frac{\|\wwh-\ww\|}{\|\ww\|}\leq \epsilon\left(\frac{\|\Xt\|\cdot\|\rr\|}{\sigma_d(\Xt)^2\|\ww\|}+\frac{\|\Xt\|}{\sigma_d(\Xt)}\right)+\epsilon^2 \frac{\|R_2\|}{\|\ww\|}.
\end{align*}
Now $\cos(\ttt)=\|\Xt\bW\ww\|/\|\bb\|$ and $\tan(\ttt)=\|\rr\|/\|\Xt\bW\ww\|$.
Plugging this into the above yields the desired result.
Again, we note this proof closely followed the analogous proof of Theorem 5.3.1 in \cite{golub2013matrix} (noting that here $f=0$ in the notation of \cite{golub2013matrix}).
\end{proof}

We next turn our attention to proving Lemma \ref{lem:2infOOS}; note we closely follow the analogous proof of Theorem 7 in \cite{levin2021limit}.
\begin{proof}
Applying Lemma 25 in \cite{levin2021limit} yields that there exists a constant $\gamma\in(0,1)$ (depending only on $F$; the key being the $\delta$-inner product property) such that with probability at least $1-D_2 n^{-2}$ ($D_2$ here depending only on $F$), 
\begin{align}
\label{eq:cos}
\cos\ttt\geq \gamma.
\end{align}
Next, following the proof of Theorem 7 in \cite{levin2021limit}, we define
$\vec{r}=\bb-\Xt\bW (\bW^T X_v)$, and we note that 
$\|\rr\|\leq\|\vec{r}\|$.
We then have that
$$\|\vec{r}^{\,T}\Xt\bW\|^2=\|\vec{r}^{\,T}\Xt\|^2=\sum_{k=1}^d\left(\sum_{j=1}^n((\bb)_j-\Xtn_j^TX_v)\Xtn_{jk}\right)^2$$
 For each $k$, conditioning on $\Xt$ and $X_v$ we note that the sum (over $j$) is the sum of $s$ mean $0$ independent random variables uniformly bounded in $[-1,1]$ (bounded here uniformly for all $\bb$) and that Hoeffding's inequality yields
 \begin{align*}
&\mathbb{P}\left(\left|\sum_{j=1}^s((\bb)_j-\Xtn_j^TX_v)\Xtn_{jk}\right|\geq 20\sqrt{s\log s}\right)\\
&=\mathbb{E}\left[\mathbb{P}\left(\left|\sum_{j=1}^s((\bb)_j-\Xtn_j^TX_v)\Xtn_{jk}\right|\geq 20\sqrt{s\log s}\,\Big|\,\Xt,X_v\right)\right]\\
&\leq \text{exp}(-5\log^2 s)\leq s^{-5}
 \end{align*}
 Taking a union over the $d$ values of $k$, we have that
 \begin{align}
 \label{eq:lastpart}
\mathbb{P}(\|\vec{r}^{\,T}\Xt\|^2\leq 400 d s\log s)\geq 1-ds^{-5}.
\end{align}

Next note that conditioning on the events in Eq. \ref{eq:2inf}, Eq. \ref{eq:evalX}, Eq. \ref{eq:cos}, and Eq. \ref{eq:lastpart} we have that
$
\nnn\leq \frac{C_2}{C_1},
$
and $\kappa_2(\Xt\bW)\leq \frac{C_2}{C_1}$ as well.
We then have 
\begin{align*}
&\frac{\|\wwh-\ww\|}{\|\ww\|}\notag\\
&\,\,\leq \frac{\|\Xhi-\Xt\bW\|}{\|\Xt\bW\|}(1+\nnn\tan\ttt)\kappa_2(\Xt\bW) +D_1\frac{dm\log(ms)^2}{s}\frac{\|\Xhi-\Xt \bW\|^2}{\|\Xt\bW\|^2\|\ww\|}\notag\\
&\leq \frac{Cm^{1/2}\log(ms)}{C_1 s^{1/2}}\left(1+\frac{C_2}{C_1\gamma}\right)\frac{C_2}{C_1}+\frac{D_1C^2dm^2\log^4(ms)}{C_1^2s^2\|\ww\|}
\end{align*}
implying that 
\begin{align}
\|\wwh-\ww\|\leq C_4\frac{Cm^{1/2}\log(ms)}{C_1 s^{1/2}}\left(1+\frac{C_2}{C_1\gamma}\right)\frac{C_2}{C_1}+\frac{D_1C^2dm^2\log^4(ms)}{C_1^2s^2}.
\label{eq:almost}
\end{align}
Following the proof of Lemma 27 in \cite{levin2021limit}, noting that 
\begin{align*}
    \|\vec r\|\geq \|\bb-\Xt\bW\ww\|=\|\Xt\bW\bW^TX_v+\vec{r}-\Xt\bW\ww\|
\end{align*}
we then have that
$$\sigma^2_d(\Xt)\|\ww-\bW^TX_v\|^2\leq\|\Xt\bW(\ww-\bW^TX_v)\|^2\leq 2(\vec{r}\,)^T\Xt\bW(\ww-\bW^TX_v)$$
from which we have (via Cauchy-Schwarz)
$$\|\ww-\bW^TX_v\|^2\leq \frac{2(\vec{r}\,)^T\Xt\bW(\ww-\bW^TX_v)}{C_1^2 s}\leq \frac{2\|(\vec{r}\,)^T\Xt\|\cdot\|\ww-\bW^TX_v\|}{C_1^2 s}$$
We then have (assuming $\|\ww-\bW^TX_v\|>0$, as equaling 0 would immediately provide the bound in Eq. \ref{eq:close}),
\begin{align}
\label{eq:close}
\|\ww-\bW^TX_v\|\leq 2\frac{\|(\vec{r}\,)^T\Xt\|}{C_1^2 s}\leq \frac{20 \sqrt{d\log s}}{C_1^2 \sqrt{s}}
\end{align}
Combining Eq. \ref{eq:close} and Eq. \ref{eq:almost} and the boundedness of $x(t)$ (and hence of $\ww$) yields the desired result.
\end{proof}

\subsubsection{Proof of Theorem \ref{thm:main}}
\label{sec:pfthm1}

By assumption [iii]. on $F$, we have that if $X,Y\stackrel{ind.}{\sim} F$, then $\mathbb{P}(\|X-Y\|\leq r)\leq c_dr^d$ (condition on $X=x$ and apply assumption [iii]. to the probability $Y$ falls into B$_r(x)$, integrating over $x$ yields the result).
We then have that ($D$ here is the constant from Lemma \ref{lem:2infOOS})
\begin{align}
\mathbb{P}&\left(\min_{v,w\in\mathcal{U}}\|X_v-X_w\|> \frac{10 D m^{1/2}\log^2(ms)}{\sqrt{s}}\right)\notag\\
&=1-\mathbb{P}\left(\exists\, v,w\in\mathcal{U}\text{ s.t. }\|X_v-X_w\|\leq \frac{10 D m^{1/2}\log^2(ms)}{\sqrt{s}}\right)\notag\\
\label{eq:ball}
&\geq 1-c_d\binom{u}{2}\frac{10^d D^d m^{d/2}\log^{2d}(ms)}{s^{d/2}}\geq 1-\xi_d u^2m^{d/2}\frac{\log^{2d}(ms) }{s^{d/2} }
\end{align}
for a suitable constant $\xi_d$ equal to (where $C$ here is independent of $d$) $\xi_d=Cc_d 10^dD^d $.
Condition now on the events in Eq. \ref{eq:2infu} and Eq. \ref{eq:ball} holding, and we see that 
\begin{itemize}
\item[1.] For all $v\in\mathcal{U}$, all OOS-embedded vertices corresponding to $X_v$ are in a ball centered at $\bW^T X_v$ of radius $\frac{Dm^{1/2}\log(ms)}{\sqrt{s}}$; this implies that 
\begin{align}
\label{eq:cluster1}
\max_{v\in\mathcal{U}}\max_{i,j\in[m]}\|\wwh-\wwhj\|\leq 2\frac{Dm^{1/2}\log(ms)}{\sqrt{s}}
\end{align}
\item[2.] For all $v,w\in\mathcal{U}$, the distance between $X_v$ and $X_w$ is at least $\frac{10 D m^{1/2}\log^2(ms)}{\sqrt{s}}$; this implies that 
\begin{align}
\label{eq:cluster12}
\min_{v,w\in\mathcal{U}, v\neq w}\min_{i,j\in[m]}\|\wwh-\wwhjw\|&\geq \min_{v,w\in\mathcal{U}, v\neq w}\|X_v-X_w\|-2\max_{v\in\mathcal{U}}\max_{i\in[m]}\|\wwh-X_v\|\notag\\
% -\min_{v,w\in\mathcal{U}, v\neq w}\min_{i,j\in[m]}\|X_w-\wwhjw\|\\
&\geq 8\frac{Dm^{1/2}\log(ms)}{\sqrt{s}}
\end{align}
\end{itemize}
From Eqs. \ref{eq:cluster1}, \ref{eq:cluster12}, the perfect matching follows immediately as the Linear Assignment Problem cost matrix $\mathbf{C}^{(i,j)}$ for matching graph $\bB^{(i)}$ to $\bB^{(j)}$ satisfies $(\bQ^{(i)})^T\mathbf{C}^{(i,j)}\bQ^{(j)}$ has diagonal values uniformly dominated by $2\frac{Dm^{1/2}\log(ms)}{\sqrt{s}}$ and off-diagonal entries uniformly lower bounded by $8\frac{Dm^{1/2}\log(ms)}{\sqrt{s}}$.  It follows that the optimal matching is given by 
\begin{align*}
\argmin_{\bQ\in\Pi_u}\text{tr}(\mathbf{C}^{(i,j)}\bQ)&=\argmin_{\bQ\in\Pi_u}\text{tr}(\bQ^{(i)}(\bQ^{(i)})^T\mathbf{C}^{(i,j)}\bQ^{(j)}(\bQ^{(j)})^T\bQ)\\
&=\argmin_{\bQ\in\Pi_u}\text{tr}((\bQ^{(i)})^T\mathbf{C}^{(i,j)}\bQ^{(j)}(\bQ^{(j)})^T\bQ\bQ^{(i)})=\{ \bQ^{(j)}(\bQ^{(i)})^T\}
\end{align*}
as desired.
\newpage

\subsection{Additional figures and simulations}
\label{app:extra}

\subsubsection{Matching accuracy simulations}

\begin{figure}[h!]
\begin{center}
\includegraphics[width=.48\textwidth]{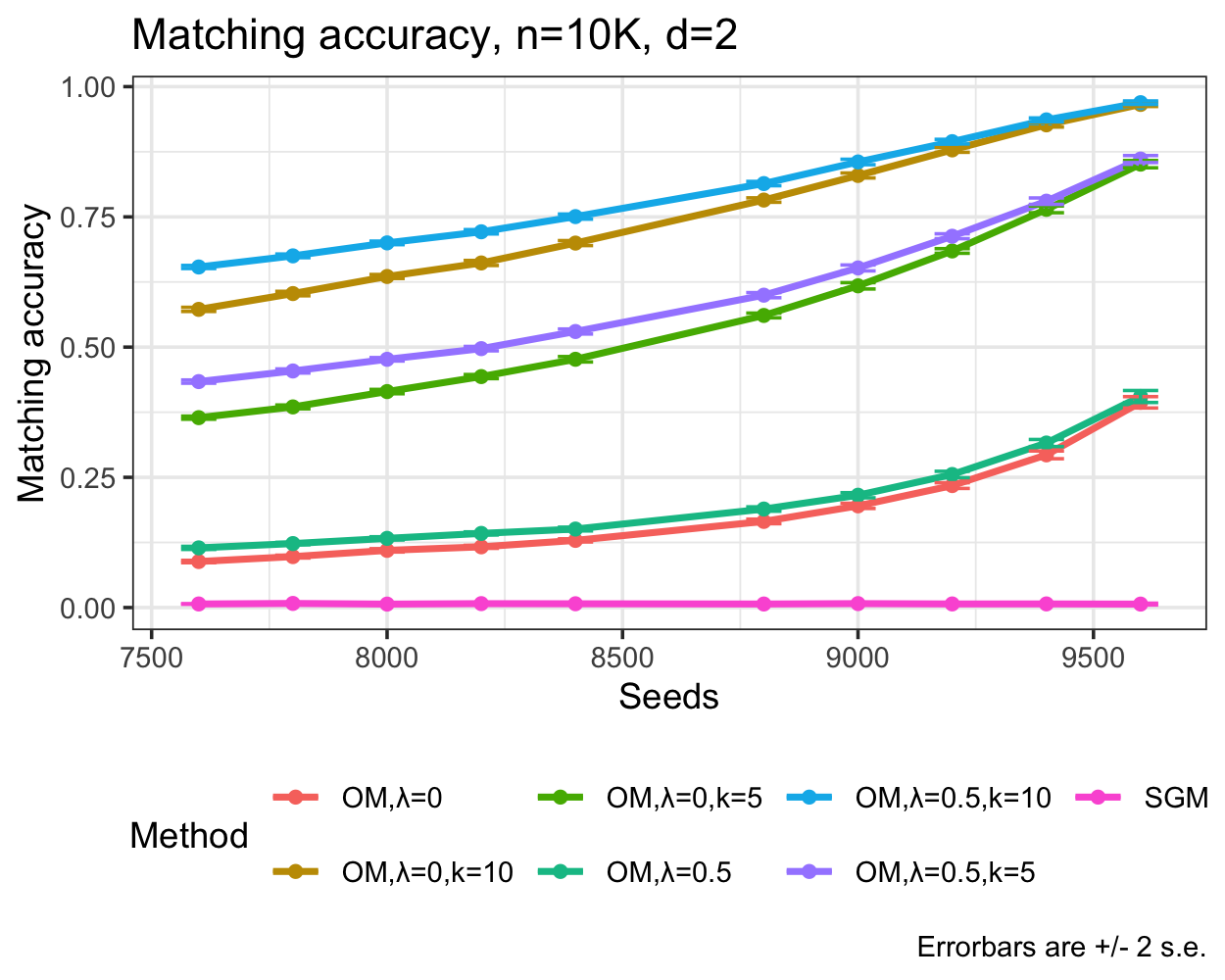}
 \includegraphics[width=.48\textwidth]{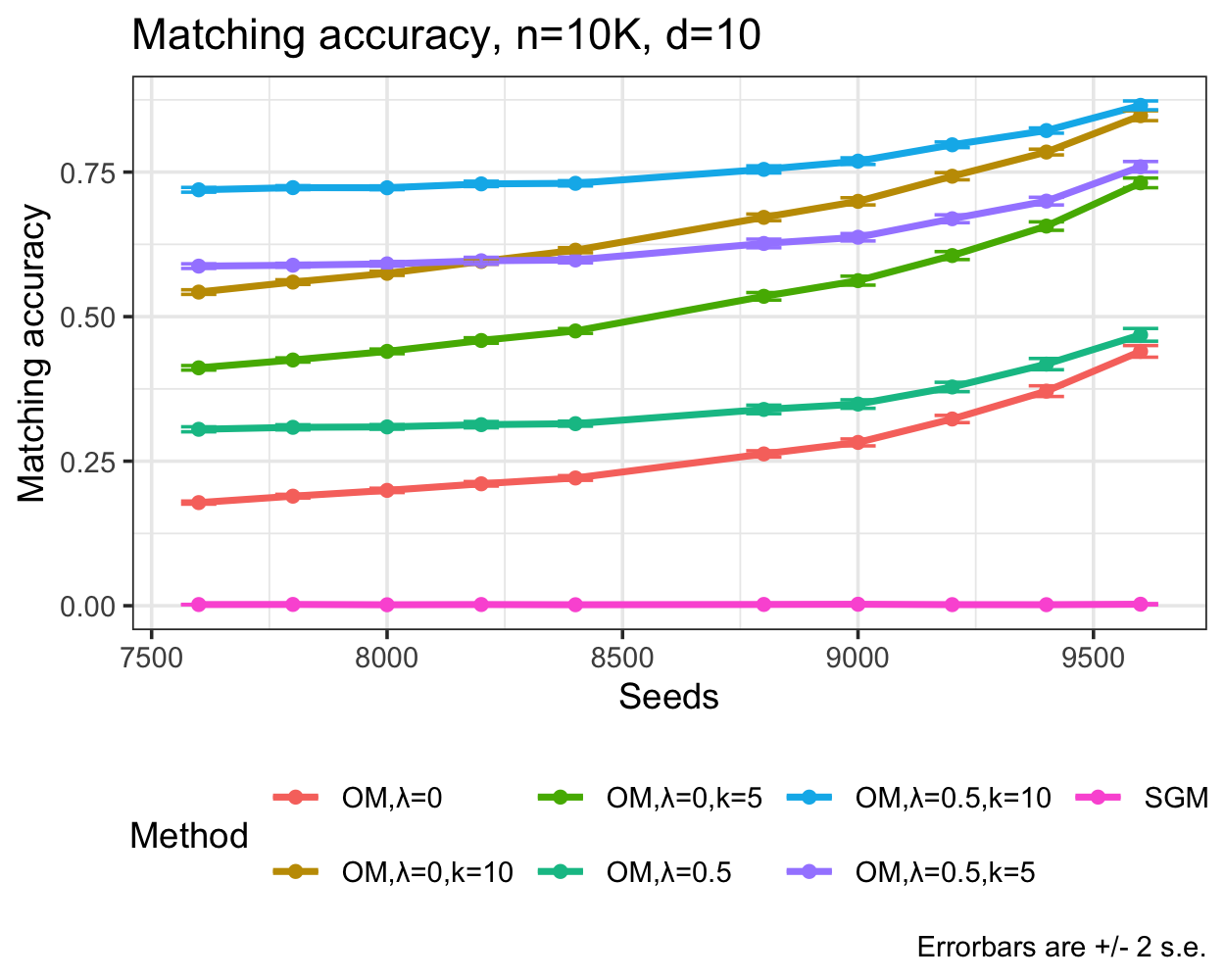}\\
\caption{The proportion of correctly unshuffled vertices for \texttt{OmniMatch}, \texttt{S-OmniMatch}, and \texttt{SGM} for a 2-d (left), and 10-d (right). Results are averaged over $nMC=30$ simulations.
In these experiments, the learning rate was lowered to $1e-6$, and the max number of iterations in gradient descent was set to 500.}
\label{plt:exp_prop_n1000}
\end{center}
\end{figure}

\begin{figure}[h!]
\begin{center}
\includegraphics[width=0.8\textwidth]{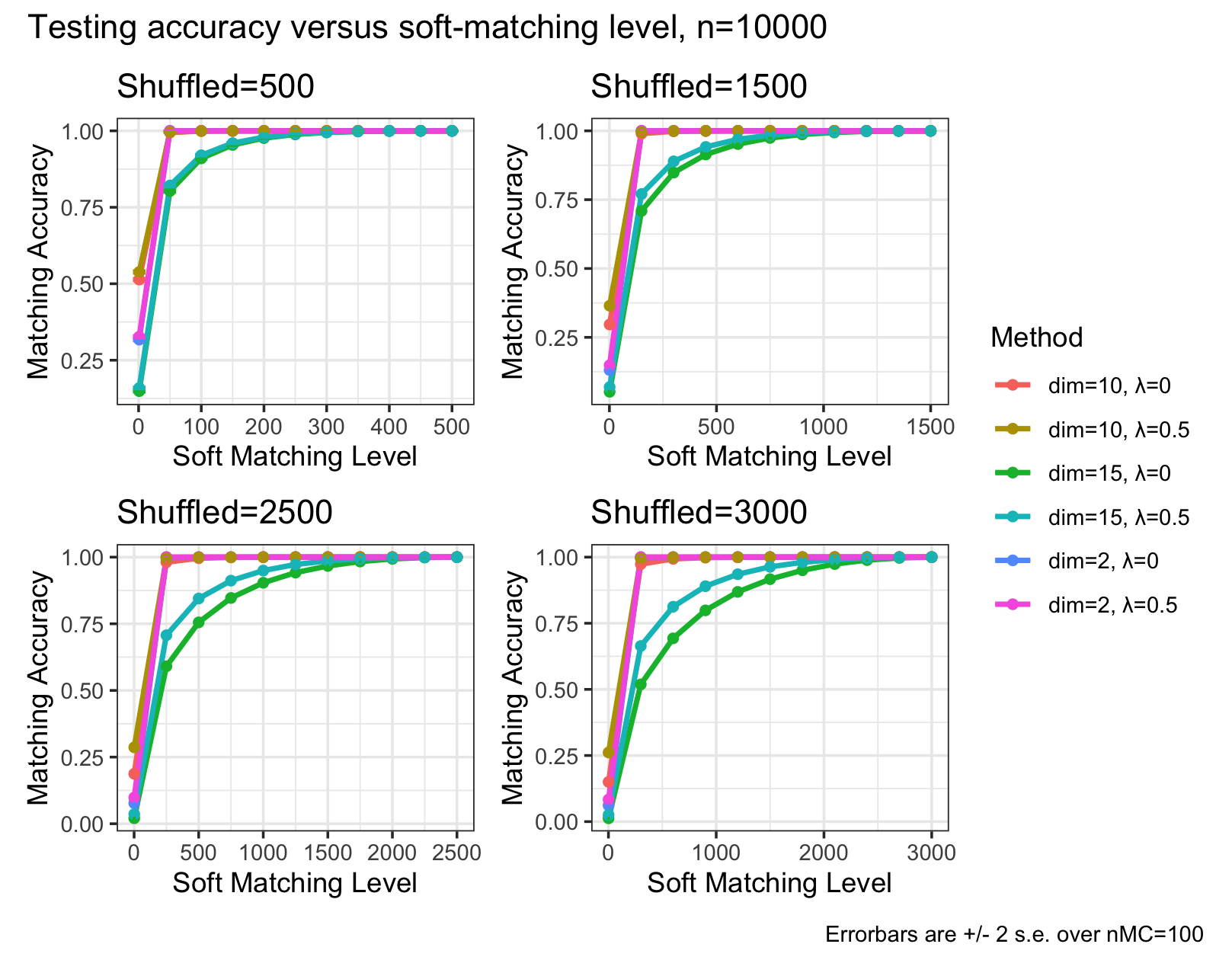}

\caption{Precision plots for the proportion of correctly unshuffled vertices with $n_s=500,\,1500,\,2500,\,3000$ shuffled vertices, where the number of neighbors used in \texttt{S-OmniMatch} ranges from 1 to $n_s$. Results are averaged over $nMC=100$ simulations for embeddings of dimensions two, ten, and fifteen of the RDPG with 10,000 vertices.}
\label{plt:exp_prop_precision_k}
\end{center} 
\end{figure}

\newpage 

\subsubsection{Multiple RDPG alignment simulations}

\begin{figure}[h!]
\begin{center}
    \includegraphics[width=.4\textwidth]{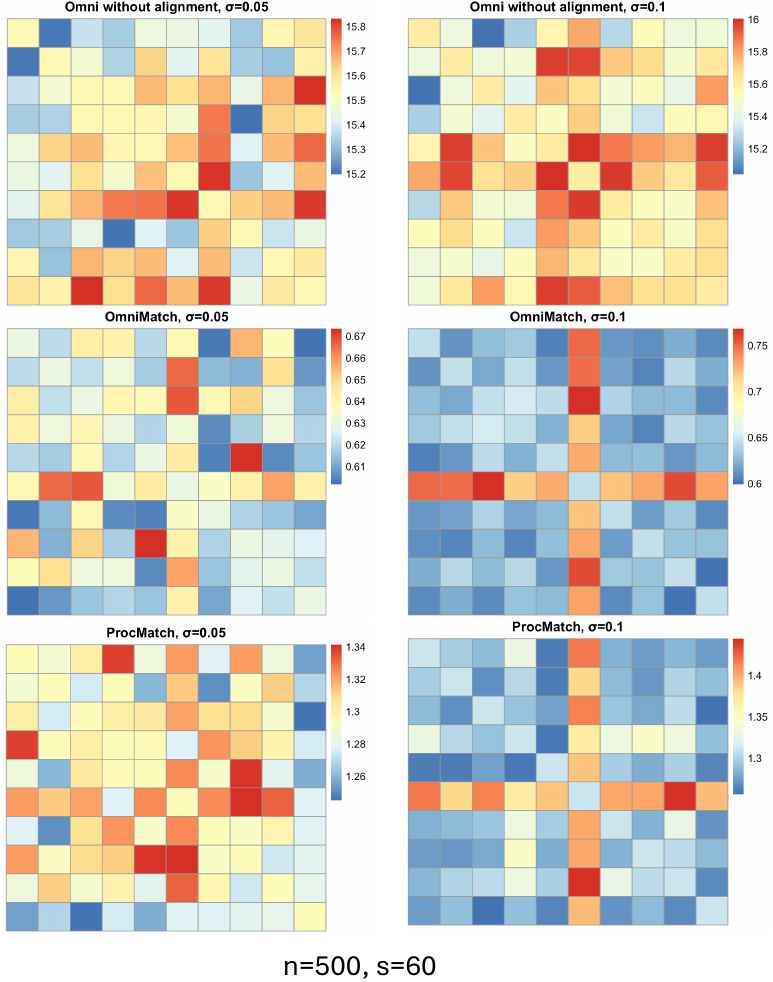}
  \caption{Using the Experimental setup as in Section 3.1.1, we plot pairwise distances among 10 RDPGs with anomalous error $\sigma=0.05$ (L) and $0.1$ (R).
  Methods employed are 
  \oo (middle row), \texttt{ProcMatch} (bottom row), and the \texttt{Omnibus} method without label matching (top row).
  Results are averaged over 30 Monte Carlo iterates.
  Note that diagonals are imputed as the average of all off-diagonal entries to allow for a more stable color gradient in the heat maps.
  }
  \label{plt:exp_multi_rdpg2}
\end{center} 
\end{figure}

\newpage
\subsection{Testing plots}
\begin{figure}[h!]
\begin{center}
\vspace{-5mm}
  \includegraphics[width=0.6\textwidth]{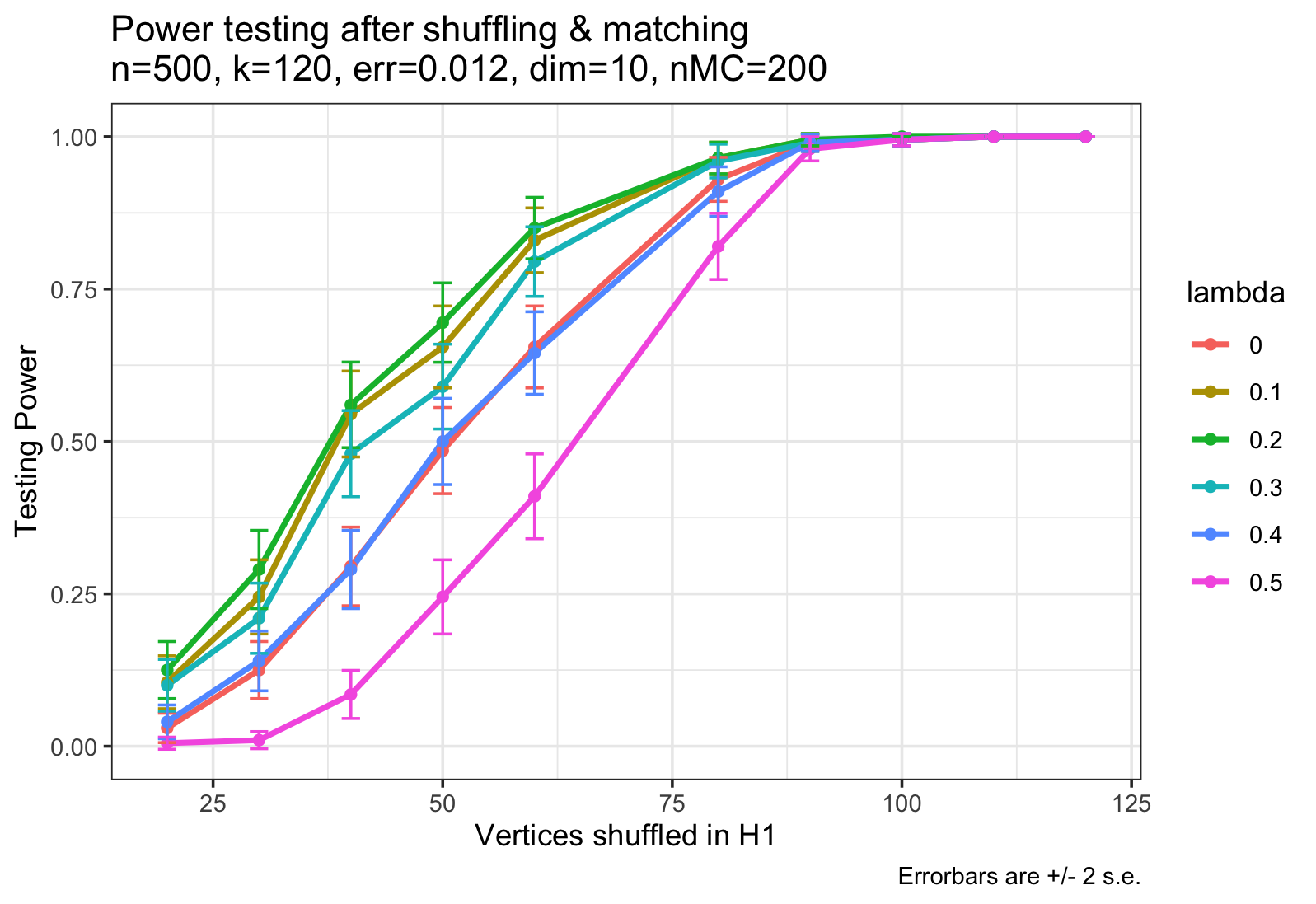}
\caption{
Statistical power for two-sample graph hypothesis testing under node shuffling; notation is as in Section \ref{sec:hypotest}. 
Here, we show the empirical testing power of \texttt{OmniMatch} as we vary $\lambda$. }
\label{plt:exp_rdpg}
\end{center}
\end{figure}

\begin{figure}[h!]
\begin{center}
\vspace{-5mm}
  %\subfloat[err$ = 0.01$]
  \includegraphics[width=1\textwidth]{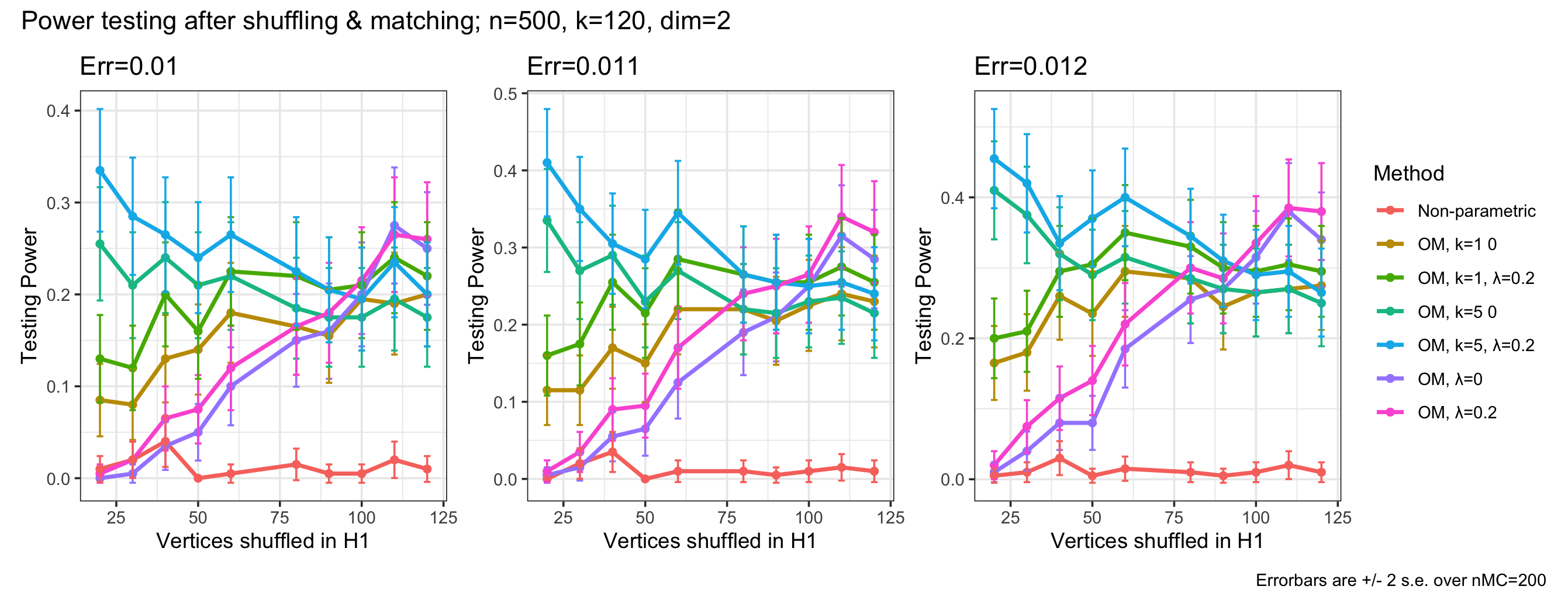}
\caption{
Statistical power for two-sample graph hypothesis testing under node shuffling. 
Panels display results from experiments on pairs of graphs with $n=500$ nodes. In each, one graph is fixed while the other has noise $\textrm{err} \in {0.01,0.011,0.012}$ added to the latent positions. All embeddings have dimension $d=2$. 
Each panel
compares the empirical testing power of \texttt{OmniMatch}, \texttt{S-OmniMatch} using 1 nearest neighbor ($k=1$), using 5 nearest neighbors ($k=5$), and a non-parametric test. }
\label{plt:exp_rdpg_d2}
\end{center}
\end{figure}

\newpage
\subsection{English-Zulu plots}
\label{sec:extraZulu}
\begin{figure}[h!]
\begin{center}
    %\subfloat[err$ = 0.012$]
    \includegraphics[width=1\textwidth]{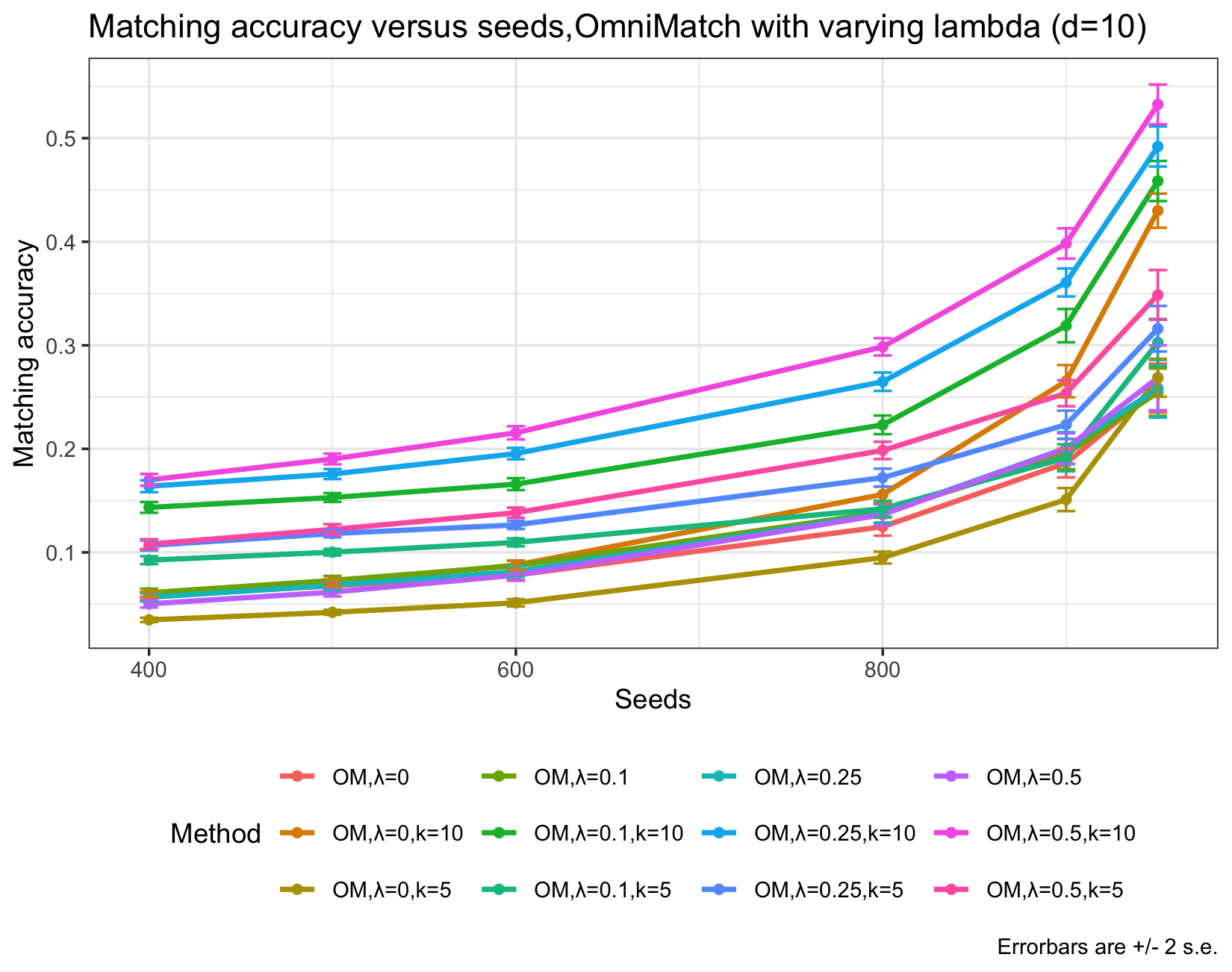} 
\caption{
Matcching accuracy for various $\lambda$ in the $d=10$ case.}
\label{fig:lls}
\end{center}
\end{figure}

\begin{figure}[h!]
\begin{center}
    %\subfloat[err$ = 0.012$]
    \includegraphics[width=01\textwidth]{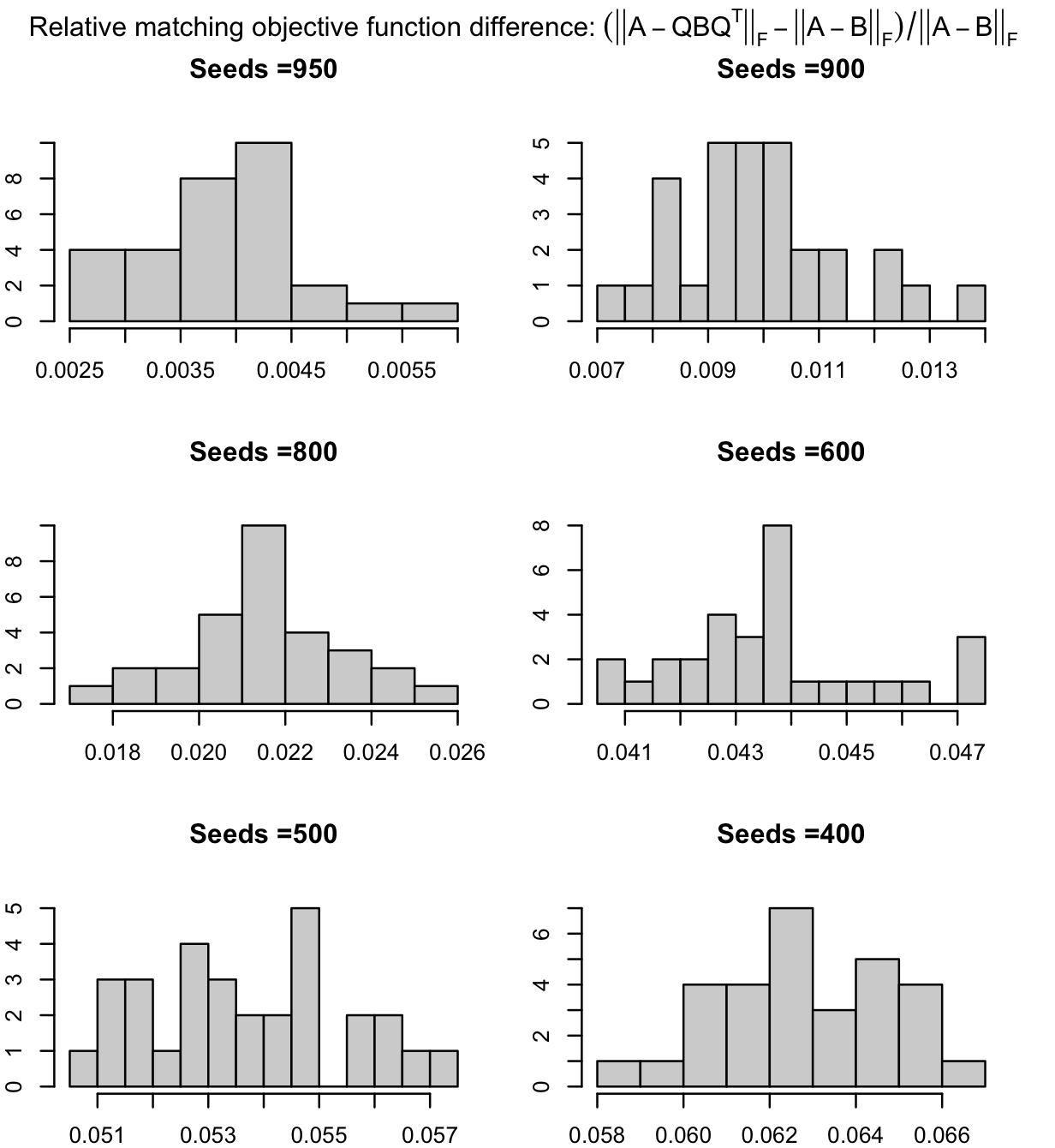} 
\caption{
For each seed set (30 randomly chosen seed sets per seed level; the same seed sets as in Section \ref{sec:engzulu}), we sample 1000 random permutations and plot a histogram of the minimum value of $\frac{\|A-QBQ^T\|_F-\|A-B\|_F}{\|A-B\|_F}$.}
\label{fig:hists}
\end{center}
\end{figure}

\end{document}